\documentclass[12pt]{article}
\usepackage{etoc}
\usepackage{titletoc}

\usepackage{graphicx,epstopdf,epsfig}
\usepackage{amsmath,amsthm,verbatim,amssymb}
\usepackage[dvipsnames, svgnames, x11names]{xcolor}
\usepackage{lscape}
\usepackage{threeparttable}
\usepackage{algorithm, algorithmic}
\usepackage{float}
\usepackage{mathrsfs}
\usepackage{booktabs}
\usepackage{multirow}
\usepackage{hyperref}
\hypersetup{colorlinks=true,
	linkcolor=blue,
	anchorcolor=blue,
	citecolor=blue}
\usepackage{enumerate}
\usepackage[round]{natbib}

\makeatletter 
\usepackage{dsfont}
\date{}
\usepackage{subcaption}

\newtheorem{theorem}{Theorem}[section]
\newtheorem{lemma}{Lemma}[section]
\newtheorem{lemma*}{Lemma}
\newtheorem{remark}{Remark}[section]
\newtheorem{remark*}{Remark}
\newtheorem{corollary}{Corollary}[section]

\newtheorem{definition}{Definition}[section]
\newtheorem{assumption}{Assumption}[section]
\usepackage{bbm}

\def\Ebb{\mathbb{E}}

\def \cA {{\cal A}}
\def \cB {{\cal B}}

\def \cD {{\cal D}}

\def \cH {{\cal H}}

\def \cL {{\cal L}}

\def \cO {{\cal O}}
\def \cP {{\cal P}}
\def \cQ {{\cal Q}}
\def \cR {{\cal R}}
\def \cS {{\cal S}}
\def \cT {{\cal T}}

\def \cW {{\cal W}}

\newcommand{\qm}{Q_{\max}}

\def \bD {\mathbb{D}}

\def \bS {\mathbb{S}}
\def \bE {\mathbb{E}}
\def \bT {\mathbb{T}}
\def \bZ {\mathbb{Z}}
\def \TV {\mathrm{TV}}

\usepackage{booktabs}
\usepackage{makecell}
\usepackage{xcolor}
\usepackage{tcolorbox}

\begin{document}

\title{Conditional Flow Models for Offline Maximum Entropy Reinforcement Learning}

\author{
Hao Cheng
\thanks{
School of Mathematics and Statistics, Wuhan University, Wuhan, 430072, China. 
Email: haocheng.math@whu.edu.cn
}
\and
Yuling Jiao
\thanks{
School of Artificial Intelligence,
 and Hubei Key Laboratory of Computational Science, Wuhan University, Wuhan, 430072,  China.
Email: yulingjiaomath@whu.edu.cn
}     
\and
Lican Kang
\thanks{Institute for Math and AI,  Hubei Key Laboratory of Computational Science, and
School of Artificial Intelligence,
Wuhan University, Wuhan, 430072, China. Email: kanglican@whu.edu.cn }
\and
Jerry Zhijian Yang
\thanks{
School of Mathematics and Statistics, and Hubei Key Laboratory of Computational Science, Wuhan University, Wuhan, 430072,  China.
Email: zjyang.math@whu.edu.cn
}
\and
Chen Zhong
\thanks{
School of Mathematics and Statistics, Wuhan University, Wuhan, 430072, China. 
Email: zcmath@whu.edu.cn
}
}

\maketitle

%%%%%%%%%%%%%%%%%%%%%%%%%%%%%%%%%%%%
%%%%%%%%%%%%%%%%%%%%%%%%%%%%%%%%%%%%
%%%%%%%%%%%%%%%%%%%%%%%%%%%%%%%%%%%%
\begin{abstract}

Offline maximum entropy reinforcement learning seeks to estimate the optimal soft action-value function \(Q^*\) from a fixed dataset without further interaction with the environment. Although \(Q^*\) is characterized by the soft optimal Bellman equation, the corresponding Bellman operator cannot be evaluated directly because both the reward distribution and the transition kernel are unknown. Consequently, estimating \(Q^*\) requires the joint treatment of environment modeling and value function learning.
%%%
We propose a framework that addresses these tasks separately. Conditional diffusion models based on probability flow ODEs estimate the conditional reward distribution and transition kernel. The resulting distributional estimators define a data driven approximation of the soft optimal Bellman operator, whose empirical Bellman residual is minimized over a class of deep neural networks to estimate \(Q^*\). This formulation separates high dimensional conditional distribution estimation from Bellman fixed point estimation while preserving the structure of the maximum entropy Bellman equation.
%%%
We establish nonasymptotic guarantees for the proposed method. For operator learning, we derive total variation error bounds for the conditional flow estimators that account explicitly for score estimation, numerical discretization, and terminal smoothing. For soft value learning, we first use the soft Bellman equation to show that $Q^*$ inherits H\"older regularity from the conditional reward and transition densities. For any H\"older index $\beta$, 
the value learning stage achieves an excess Bellman residual risk bound  of order $\widetilde{\mathcal O}\big(n^{-\frac{2\beta}{d+2\beta}}\big),$ where $d$ is the total dimension of the state and action spaces.
In contrast to many existing theoretical analyses of deep reinforcement learning, our results do not require Bellman completeness. Numerical experiments support the theoretical findings and demonstrate the effectiveness of the proposed method.

\vspace{0.5cm} 
\noindent{\bf KEY WORDS}: 
Deep maximum entropy RL,
Bellman residual minimization,
Deep $Q^\ast$ estimation, Conditional flow model.

\end{abstract}

\section{Introduction}
Reinforcement learning (RL) studies sequential decision making in Markov decision processes, where an agent learns from observations, actions, and rewards to improve long-term performance \citep{kaelbling1996reinforcement,sutton2018reinforcement}. Classical methods represent value functions or policies explicitly, whereas deep reinforcement learning uses neural networks to handle large or continuous state and action spaces. This combination has enabled progress in visual control and game playing \citep{mnih2015human,silver2016mastering}, as well as applications to chip placement and plasma control \citep{mirhoseini2021graph,degrave2022magnetic}. These applications illustrate the flexibility of neural representations, while also making the statistical behavior of value estimation an important theoretical question. In value-based methods, the fitted target contains a continuation value estimated from data. Consequently, errors in Bellman-target construction interact with neural network approximation and empirical optimization. Actor-critic methods introduce a further interaction because policy updates depend on an estimated critic, while the critic is trained using data and targets associated with the policy \citep{fujimoto2018,haarnoja2018soft}. In the offline setting, the available dataset is fixed and additional environment interaction is unavailable. A useful analysis must explain how offline data are used to estimate the Bellman operator and how operator estimation error propagates to the learned value function.

Maximum entropy RL augments the expected-return objective with policy entropy, encouraging stochastic policies while retaining a preference for actions with high expected return \citep{ziebart2008maximum,haarnoja2018soft,levine2018reinforcement}. Entropy regularization replaces action maximization in the optimal Bellman equation with a log-partition operation. The resulting operator is associated with a Gibbs policy whose density depends on the soft action-value function \citep{geist2019theory}. This structure connects value function estimation with policy optimization: the soft value summarizes an integral over actions, and the estimated action-value function determines the corresponding policy. Soft Q-learning develops this connection through energy-based policies and approximate inference \citep{haarnoja2017reinforcement,levine2018reinforcement}. Path Consistency Learning uses temporal consistency relations involving values and entropy-regularized action probabilities \citep{nachum2017bridging}, while Soft Actor-Critic (SAC) combines a stochastic actor, a soft critic, and replay data for continuous control \citep{haarnoja2018soft,haarnoja2018sacapps}. The relationship between policy gradients and soft Q-learning provides a complementary account of their shared optimality structure \citep{schulman2017equivalence}. Soft policy-gradient methods and robust-control analyses further develop the algorithmic and theoretical consequences of entropy regularization \citep{shi2019soft,eysenbach2022maximum}. For offline estimation, however, the soft Bellman operator still involves a conditional expectation over the unknown next-state distribution. Evaluating this expectation, together with the action integral, is the statistical and computational problem addressed by the present construction.

Model-based RL provides a natural way to address this difficulty by explicitly learning the reward or transition dynamics from data and then using the learned model for planning, synthetic data generation, or Bellman evaluation \citep{sutton2018reinforcement,janner2019trust,yu2020mopo,kidambi2020}. This paradigm is particularly attractive when real environment interaction is expensive or unavailable, since the learned dynamics can be queried repeatedly. Its central challenge is model bias: inaccuracies in the learned dynamics may accumulate over long imagined rollouts and can be amplified by subsequent policy optimization. Existing approaches such as MBPO, MOPO, and MOReL mitigate this effect through short model rollouts or conservative and uncertainty-aware objectives \citep{janner2019trust,yu2020mopo,kidambi2020}. This perspective suggests learning the conditional next state law directly and using it locally to approximate the conditional expectation appearing in the soft Bellman operator.

Recent advances in generative flow models have provided flexible tools for representing complex distributions in reinforcement learning. Existing flow based reinforcement learning methods mainly use flows on the decision side to model behavior policies, represent multimodal action distributions, or generate high value actions. For example, Flow Q Learning uses a flow matching policy to capture complex behavior distributions and introduces an auxiliary one step policy to avoid backpropagation through the full iterative sampling procedure \citep{park2025}. Energy guided and value guided flow methods further combine generative flow sampling with value functions or energy signals to improve action generation and policy optimization \citep{alles2025,zhang2025,feng2025}. These studies demonstrate the strong distributional modeling capability of flow models in reinforcement learning. However, this capability can also be used on the environment side. In particular, a conditional flow model can approximate the full transition law $P(S' |  s,a)$ rather than only predicting the conditional mean of the next state. Repeated samples generated from the learned transition law can then be used to approximate the conditional expectation appearing in the soft Bellman operator.

Motivated by these developments, we propose a framework for offline maximum entropy value learning that integrates conditional probability flow modeling with soft Bellman residual risk. Conditional probability flow ODEs estimate the conditional reward law and transition law from batch data. The learned reward and transition distributions are incorporated into the soft optimal Bellman operator. With the learned environment model held fixed, the optimal soft action value function $Q^*$ is estimated by minimizing the empirical soft Bellman residual over a ReLU DNN class. Unlike existing flow based reinforcement learning methods that mainly use flow models to represent policies or generate actions, our flow models operate on the environment side and model both the reward and the next state conditional on $(s,a)$. This construction separates environment law estimation from soft Bellman residual risk, enables repeated sampling from the learned reward and transition distributions, and avoids jointly optimizing the environment model and the value function within a single coupled objective.
%%%
We derive a nonasymptotic bound on the squared Bellman-residual risk. The bound separates the error in the learned operator from the statistical and approximation terms associated with residual minimization. Under the stated conditions, the latter terms are of order \(\widetilde O(n^{-2\beta/(d+2\beta)})\), where \(n\) is the value learning sample size, \(d\) is the state-action dimension, and \(\beta\) is the Hölder smoothness of \(Q^*\).
%%%
Importantly, our analysis does not require
the Bellman completeness assumption commonly adopted in theoretical studies
of deep reinforcement learning
\citep{uehara2023offline,zhan2022offline}.
%%%
Empirically, we evaluate the proposed method in three experimental settings: a simulated MDP, classical control tasks, and HalfCheetah datasets from D4RL \citep{fu2020d4rl}. We compare the proposed method with  SAC \citep{haarnoja2018sac}, CQL \citep{kumar2020conservative}, and MOPO \citep{yu2020mopo}. The experimental results demonstrate the effectiveness of using conditional probability flow models to estimate the reward and transition laws and support the separation of environment law estimation from soft value function learning.

\subsection{Contributions}
The main contributions of this work are summarized as follows.

\begin{itemize}
\item 
We develop a framework for offline maximum entropy RL that separates conditional environment modeling from soft value learning. Conditional probability flow ODEs learn the reward distribution and transition kernel, and samples generated from these models define a data-driven approximation of the soft optimal Bellman operator. We then estimate $Q^{\ast}$ by minimizing the empirical soft Bellman residual over a class of ReLU DNNs. This formulation allows the environment modeling and value learning errors to be analyzed separately. Numerical results show that the proposed method achieves competitive performance across the considered offline control tasks.
\item 
We decompose the error of the proposed estimator into operator estimation and value learning terms. The operator error is controlled by total variation bounds for the learned reward and transition laws. We further establish that the H\"older regularity of the conditional densities with respect to the state-action variables is inherited by $Q^*$ through the soft Bellman equation. This result yields rates for the value learning stage without a separate smoothness assumption on $Q^*$ or a Bellman completeness condition. Combining these results, we obtain nonasymptotic bounds on the operator estimation error and the Bellman residual risk of the resulting action-value estimator.
\end{itemize}

\subsection{Related Work}
\paragraph{Deep RL and Max Entropy RL.}
 DRL combines sequential decision making with neural function approximation, allowing value functions and policies to be learned from high dimensional observations \citep{sutton2018reinforcement,mnih2015human}. Its main difficulty is that the learning target is endogenous: a value estimate is reused to construct later temporal difference targets, while a policy update changes the distribution from which future data are collected \citep{sutton2018reinforcement,jiao2025deep}. DRL algorithms therefore differ largely in how they control  errors, distribution shift, and the interaction between policy evaluation and policy improvement \citep{sutton2018reinforcement,mnih2015human,jiao2025deep}. In discrete action spaces, DQN established a practical value-based template by combining a convolutional Q-network with experience replay and a slowly updated target network \citep{mnih2015human}. Replay improves data reuse and weakens short range temporal dependence, whereas the target network slows the change of the Bellman target \citep{mnih2015human}. Double DQN reduces maximization bias by separating action selection from action evaluation \citep{vanhasselt2016}; the dueling architecture decomposes the output into state value and advantage streams \citep{wangdueling2016}; and distributional RL models the return distribution rather than only its expectation \citep{bellemare2017}. These refinements improve target construction or representation, but the learned Q-function is still defined through a recursive Bellman relation and remains sensitive to approximation error in weakly sampled regions \citep{sutton2018reinforcement,fujimoto2018,jiao2025deep}.

Maximum entropy RL modifies this conventional objective by maximizing expected return together with the policy entropy. The resulting soft Bellman backup replaces action maximization by a log-sum-exp (or log partition) operation, and its optimal policy is the corresponding Gibbs distribution \citep{ziebart2008maximum,levine2018reinforcement,geist2019theory}. This formulation preserves multiple competitive actions, which can improve exploration and robustness when the value landscape is multimodal. In deep continuous-control settings, soft Q-learning introduced energy-based stochastic policies \citep{haarnoja2017reinforcement}, while SAC made the maximum entropy objective practical through off policy replay, separate actor and critic networks, and an adaptive temperature variant \citep{haarnoja2018soft,haarnoja2018sacapps}. Path Consistency Learning offers a complementary view by enforcing multi-step soft consistency between values and entropy regularized policy probabilities \citep{nachum2017bridging}. These methods differ in parameterization and optimization, but all use entropy regularization to couple policy improvement with soft value estimation.

Further studies have expanded both the algorithmic and theoretical foundations of learning with entropy regularization. The equivalence between policy gradient methods and soft Q-learning reveals that maximum entropy methods based on value functions and those based on policies share the same underlying optimization structure \citep{schulman2017equivalence}. Deep Soft Policy Gradient provides an alternative implementation of the maximum entropy objective in continuous action spaces by combining actor and critic networks with off-policy learning \citep{shi2019soft}. For offline learning, minimax soft-Q learning provides guarantees under realizability and partial coverage \citep{uehara2023offline}; separately, robust control results show that MaxEnt RL can solve classes of RL problems with reward and transition uncertainty \citep{eysenbach2022maximum}. Together, these studies connect entropy regularization to algorithm design and statistical guarantees, while leaving open the question of how transition model error affects soft Bellman value learning.

Continuous control makes explicit maximization over actions impractical and motivates policy gradient and actor--critic methods \citep{konda1999,silver2014}. In an actor--critic architecture, the actor represents the policy, while the critic estimates a value or advantage used to reduce gradient variance \citep{konda1999,mnih2016}. This two line organization is flexible, but it creates coupled optimization: the critic is fitted using data from a changing actor, and the actor is updated using a critic that is only approximately accurate \citep{konda1999,fujimoto2018}. Deterministic policy gradient methods differentiate the critic with respect to the action and backpropagate this derivative through the actor \citep{silver2014}. DDPG combines this mechanism with replay and target networks to obtain an off policy algorithm for continuous actions \citep{lillicrap2015}. Its deterministic actor is efficient, but exploration must be injected separately and the actor may exploit local overestimation in the critic \citep{lillicrap2015,fujimoto2018}. TD3 addresses this failure mode through clipped double-Q targets, delayed actor updates, and target policy smoothing \citep{fujimoto2018}. These changes improve robustness without removing the fundamental dependence of policy improvement on an estimated critic \citep{fujimoto2018}.

Explicit environment modeling is central to our setting. Direct DRL methods construct Bellman targets from observed transitions without first estimating a separate transition kernel, whereas model-based methods estimate rewards or dynamics and use the learned model for planning, synthetic data generation, or Bellman evaluation \citep{sutton2018reinforcement,janner2019trust,yu2020mopo,kidambi2020}. Explicit modeling can improve data reuse, but inaccurate models may be exploited by the policy and multi-step rollouts can accumulate error \citep{janner2019trust,yu2020mopo,kidambi2020}. MBPO limits this effect with short rollouts initialized from real data \citep{janner2019trust}; MOPO and MOReL introduce pessimism or uncertainty control in offline model-based learning \citep{yu2020mopo,kidambi2020}. Our framework learns the conditional next state law, but uses it only for the one step expectation in the soft Bellman operator rather than for long imagined trajectories. This localized construction makes transition error visible at the operator level and separates it from the statistical and approximation errors of Bellman residual value estimation, in contrast to the repeated model rollout used in MBPO, MOPO, and MOReL \citep{janner2019trust,yu2020mopo,kidambi2020}.

\paragraph{Flow-based RL.}
Generative modeling has broadened the policy and trajectory distributions available to RL algorithms \citep{janner2022,wang2023,chi2023,park2025}. A conventional continuous control actor is often a deterministic map or a diagonal Gaussian distribution \citep{lillicrap2015,haarnoja2018soft}. Although computationally convenient, such a representation can be inadequate when a state admits several qualitatively different yet individually valid actions; a unimodal actor may average incompatible modes and place mass on actions that are poorly supported by the data \citep{wang2023,hansenestruch2023,chi2023,celik2025dime,dong2025maximum}. Diffusion and flow matching models instead transport a simple reference distribution into a complex conditional distribution, making multimodal action sequences, policies, or trajectories directly representable \citep{song2021,lipman2023,albergo2025stochastic}. This flexibility is particularly attractive in offline RL, goal conditioned control, imitation learning, and robotic manipulation, where datasets may contain several successful behavioral strategies \citep{janner2022,wang2023,hansenestruch2023,chi2023,park2025}.

For flow-based generative policies in particular, direct policy improvement through an iterative flow remains difficult because the terminal action depends on every integration step \citep{park2025,alles2025}. Flow Q-Learning (FQL) separates distribution modeling from Q-based improvement \citep{park2025}. It first fits a flow matching policy to a potentially multimodal behavior distribution, then trains an auxiliary one-step policy with a Q-learning objective and links the two through a distillation style loss \citep{park2025}. The construction avoids backpropagating the RL objective through the complete flow trajectory and removes iterative sampling at test time \citep{park2025}. The flow supplies an expressive behavioral prior, whereas the one-step policy provides a tractable object for value improvement \citep{park2025}. Importantly, the flow in FQL models the conditional action distribution and is therefore a policy generator, not an estimator of the environment transition law \citep{park2025}.

A second group of methods incorporates energy or value information directly into the probability path. FlowQ absorbs energy guidance into the learned velocity field rather than repeatedly applying an external guidance term during inference \citep{alles2025}. Energy weighted flow matching reweights the flow objective so that the terminal distribution approximates an energy tilted reference distribution, with a learned Q-function providing the energy in offline RL \citep{zhang2025}. General analyses of flow guidance describe how conditional expectations, energy gradients, and approximations of intermediate guided marginals determine the corrected velocity field \citep{feng2025}. Diffusion guidance can likewise be interpreted as a controllable policy improvement operator, where the guidance scale balances fidelity to a reference generator against movement toward critic preferred actions \citep{frans2025diffusion}. Maximum entropy approaches such as DIME and MaxEntDP use expressive diffusion policies to retain multiple high value action modes rather than collapsing to a single deterministic decision \citep{celik2025dime,dong2025maximum}.

These methods primarily improve the representation of the policy $\pi(a|  s)$ \citep{wang2023,hansenestruch2023,park2025,alles2025,zhang2025,celik2025dime,dong2025maximum}. They do not, by themselves, identify the environmental transition law $P(s'|  s,a)$, because the generators in these methods are trained as action or trajectory models rather than conditional next state models \citep{janner2022,wang2023,hansenestruch2023,chi2023,park2025,alles2025}. This distinction is essential for maximum entropy value learning because the soft Bellman target contains both an action side log partition operation and an expectation over the next state distribution \citep{haarnoja2018soft,geist2019theory}. An action flow can generate diverse actions, but without environment interaction or a separate dynamics model it cannot generate corresponding next states \citep{park2025,alles2025}; conversely, a learned transition model can sample next states conditional on $(s,a)$ but does not automatically define an improved policy \citep{janner2019trust,yu2020mopo,kidambi2020}. Policy flow methods therefore address multimodal action representation and value guidance, whereas our method addresses conditional transition estimation and the effect of transition error on the soft Bellman operator \citep{wang2023,park2025,alles2025,janner2019trust,yu2020mopo,kidambi2020}.
In contrast, our method uses flows to learn the reward and transition distributions, which are then used to construct an empirical Bellman operator for value function learning.

\section{Preliminaries}
%%%%%%%%%%%%%%%%%%%%%%%%%%%%%
\subsection{Markov Decision Process }
A discounted Markov decision process (MDP) is described by $(\cS,\cA,P,\cR,\gamma)$. Here, $\cS$ and $\cA$ are the state and action spaces, $\gamma\in[0,1)$ is the discount factor, and $\cR(\cdot|s,a)$ is the conditional distribution of the immediate reward $R(s,a)$. The transition kernel $P(\cdot|s,a)$ gives the distribution of the next state after taking action $a$ in state $s$.
Let $\cB(\cS)$ denote the Borel $\sigma$-algebra on $\cS$, and let $M(\cS)$ denote the set of probability measures on $(\cS,\cB(\cS))$. The kernel $P:\cS\times\cA\to M(\cS)$ satisfies the following measurability condition: for every $D\in\cB(\cS)$, the map $(s,a)\mapsto P(D|s,a)$ is measurable. A policy $\pi$ specifies the action distribution $\pi(\cdot|s)$ at each state $s$.
Given an initial state distribution $P_1$ and a policy $\pi$, a trajectory $\{S_i,A_i,R_i,S_i^{\prime}\}_{i=1}^n$ is generated as follows, where $S_i^{\prime}:=S_{i+1}$:
\[
S_1\sim P_1,~~
A_i\sim\pi(\cdot|  S_i),~~
R_i\sim\cR(\cdot|  S_i,A_i),~~
S_i^{\prime}\sim P(\cdot|  S_i,A_i),~~i=1,\ldots,n.
\]
For a state-action pair (s, a) and a given policy $\pi$, the action-value function is defined by
\[
Q^\pi(s,a):=\Ebb \Big[\sum_{t=0}^{\infty}\gamma^t R_{t+1}|  S_1=s,  A_1=a,  \pi\Big].
\]
Suppose that rewards satisfy $R(s,a)\in[0,R_{\max}]$ for every $(s,a)\in\cS\times\cA$. Then $Q^\pi$ takes values in $[0,R_{\max}/(1-\gamma)]$.
Assume that an optimal policy $\pi^*$ exists, and write $Q^*:=Q^{\pi^*}$. Then $Q^*$ satisfies the Bellman optimality equation $Q^*=\mathcal T^*Q^*$,
where the optimal Bellman operator $\mathcal T^*$ is defined as
\begin{equation*}  
	\mathcal{T}^* Q(S,A) = \mathbb{E}[R(S,A)] + \gamma \mathbb{E}_{S' \sim P(\cdot | S,A)}\max_{a^{\prime} \in \mathcal{A}} Q(S', a^{\prime}).
\end{equation*}

\subsection{Entropy-Regularized Markov Decision Processes}\label{sec:entropy-mdp}
The Markov decision process introduced above evaluates a policy by its expected discounted cumulative reward. Max entropy reinforcement learning augments this objective with an entropy regularization term, thereby encouraging stochastic policies and exploration
\citep{ziebart2008maximum,haarnoja2018soft,levine2018reinforcement,geist2019theory}. For a policy density $\pi(\cdot|  s)$, its entropy at state $s$ is defined as
\[
\mathcal H\big(\pi(\cdot|  s)\big):=-\mathbb E_{A\sim\pi(\cdot|  s)}\big[\log\pi(A|  s)\big].
\]
The temperature parameter $\alpha>0$ determines the relative contribution of the entropy term. The resulting entropy regularized objective is
\[
J(\pi):=\mathbb E^\pi_{S_1\sim P_1}\left[\sum_{t=1}^{\infty}\gamma^{t-1}\Big(R_t+\alpha\mathcal H\big(\pi(\cdot|  S_t)\big)\Big)\right].
\]
Here, the expectation is taken over the initial state $S_1\sim P_1$ and the trajectory generated by $\pi$, including the initial action $A_1\sim\pi(\cdot|S_1)$. Maximum entropy reinforcement learning seeks to maximize $J(\pi)$ over admissible policies.
Throughout the paper, we assume that the policy entropy is uniformly bounded, namely, $\big|\mathcal H(\pi(\cdot|  s))\big|\le H_{\max}<\infty$ for every admissible policy $\pi$ and state $s$. For a fixed policy $\pi$, we define the soft action value and state value functions through
\begin{gather*}
Q^\pi(s,a)=\mathbb E_{R\sim\mathcal R(\cdot|  s,a)}[R]+\gamma\mathbb E_{S'\sim\mathcal P(\cdot|  s,a)}\big[V^\pi(S')\big],\\
V^\pi(s)=\mathbb E_{A\sim\pi(\cdot|  s)}\big[Q^\pi(s,A)-\alpha\log\pi(A|  s)\big].
\end{gather*}
Under this convention, $Q^\pi$ includes the current reward and all future rewards and entropy bonuses, while the current entropy bonus is included in $V^\pi$. Consequently, $J(\pi)=\mathbb E_{S_1\sim P_1}[V^\pi(S_1)]$.
Equivalently, $Q^\pi$ is the fixed point of the soft Bellman operator
\[
\mathcal T_\alpha^\pi Q(s,a):=\mathbb E_{R\sim\mathcal R(\cdot|  s,a)}[R]+\gamma\mathbb E_{S'\sim\mathcal P(\cdot|  s,a)}\mathbb E_{A'\sim\pi(\cdot|  S')}\big[Q(S',A')-\alpha\log\pi(A'|  S')\big].
\]
For any bounded function $Q$, the Gibbs variational principle gives
\[
V(s):=\max_{\pi(\cdot|  s)}\mathbb E_{A\sim\pi(\cdot|  s)}\big[Q(s,A)-\alpha\log\pi(A|  s)\big]=\alpha\log\int_{\mathcal A}\exp\left(\frac{Q(s,a)}{\alpha}\right)da,
\]
and the corresponding maximizing policy is
\[
\pi_Q(a|  s)=\frac{\exp\big(\frac{Q(s,a)}{\alpha}\big)}{Z_Q(s)},~~ Z_Q(s)=\int_{\mathcal A}\exp\left(\frac{Q(s,\widetilde a)}{\alpha}\right)d\widetilde a.
\]
The soft optimal Bellman operator can therefore be written as
\[
\mathcal T_\alpha^*Q(s,a):=\mathbb E_{R\sim\mathcal R(\cdot|  s,a)}[R]+\gamma\alpha\mathbb E_{S'\sim\mathcal P(\cdot|  s,a)}\left[\log\int_{\mathcal A}\exp\left(\frac{Q(S',a')}{\alpha}\right)da'\right].
\]
The optimal soft action-value function is the fixed point of this operator:
\[
Q^*(s,a)=\mathcal T_\alpha^*Q^*(s,a)=\mathbb E_{R\sim\mathcal R(\cdot|  s,a)}[R]+\gamma\alpha\mathbb E_{S'\sim\mathcal P(\cdot|  s,a)}\left[\log\int_{\mathcal A}\exp\left(\frac{Q^*(S',a')}{\alpha}\right)da'\right].
\]
At the fixed point $Q^*$, the associated optimal state value function and policy are
\begin{gather*}
V^*(s)=\alpha\log\int_{\mathcal A}\exp\left(\frac{Q^*(s,a)}{\alpha}\right)da,\\
\pi^*(a|  s)=\frac{\exp\big(\frac{Q^*(s,a)}{\alpha}\big)}{Z_Q^*(s)},~~Z_Q^*(s)=\int_{\mathcal A}\exp\left(\frac{Q^*(s,\widetilde a)}{\alpha}\right)d\widetilde a. 
\end{gather*}
Under the bounded reward condition $0\le R\le R_{\max}$ introduced in the preceding section and the entropy bound $\big|\mathcal H(\pi(\cdot|  s))\big|\le H_{\max}$, the soft action value functions are uniformly bounded:
\[
|Q^\pi(s,a)|\le\sum_{t=0}^{\infty}\gamma^t\big(R_{\max}+\alpha H_{\max}\big)=\frac{R_{\max}+\alpha H_{\max}}{1-\gamma}=:Q_{\max}.
\]
In this paper, we consider three mutually independent batch datasets collected under a behavior policy $\pi_b$:
$$
\mathbb D_r:=\{(S_i,A_i,R_i,S_i')\}_{i=1}^{n_r},~~ \mathbb D_x:=\{(S_i,A_i,X_i)\}_{i=1}^{n_x},~~ \mathbb S:=\{(S_i,A_i,R_i,S_i')\}_{i=1}^{n}.
$$
Here, $X_i:=S_i'$ denotes the next state observation associated with $(S_i,A_i)$, while $n_r$, $n_x$, and $n$ are the sample sizes of $\mathbb D_r$, $\mathbb D_x$, and $\mathbb S$, respectively. The observations within each dataset are assumed to be independently and identically distributed. We use $\mu_{\pi_b}$, abbreviated as $\mu$, to denote the joint distribution of the random state action pair $(S,A)$ induced by $\pi_b$. 
We consider only the intersection of the supports of $\mu_{\pi_b}$ and $\mu_{\pi^*}$ and, without loss of generality, take this common support to be $\mathcal S\times\mathcal A=[0,1]^{d_s}\times[0,1]^{d_a}$. Here, $\mu_{\pi^*}$ denotes the joint state-action distribution induced by the optimal policy $\pi^*$.
The dataset $\mathbb D_r$ is used to estimate the conditional reward distribution $\mathcal R(\cdot|  s,a)$, whereas $\mathbb D_x$ is used to estimate the conditional transition distribution $\mathcal P(\cdot|  s,a)$, denoted by $p_1(\cdot|  s,a)$ in the conditional flow construction. The independent dataset $\mathbb S$ is used for value function learning. The corresponding estimation procedures are presented in Section~\ref{method}.

\subsection{Probability Flow ODEs for Score-Based Generative Models} \label{sec:score-flow}
Let \(\{X_t\}_{t\in[0,1]}\) be a stochastic path on \(\mathbb R^d\), and let \(p_t\) denote the marginal density of \(X_t\). Assume that the path is differentiable at interior times and satisfies the integrability conditions needed to differentiate expectations of smooth, compactly supported test functions. Define the marginal velocity by \(u^*(x,t)=\mathbb E[\dot X_t|  X_t=x]\). Within the framework of stochastic interpolants, \(p_t\) satisfies the following continuity equation in the weak sense \citep{albergo2025stochastic}
\[
\partial_t p_t(x)+\nabla_x\cdot\big(u^*(x,t)p_t(x)\big)=0.
\]
For any $\varepsilon(t)\ge0$, this marginal path can also be realized by the SDE
\[
dZ_t=\left(u^*(Z_t,t)+\varepsilon(t)\nabla_x\log p_t(Z_t)\right)dt+\sqrt{2\varepsilon(t)} dB_t .
\]
Its Fokker--Planck equation is
\[
\partial_t p_t(x)=-\nabla_x\cdot\left[\left(u^*(x,t)+\varepsilon(t)\nabla_x\log p_t(x)\right)p_t(x)\right]+\varepsilon(t)\Delta_xp_t(x),
\]
which reduces to $\partial_t p_t(x)+\nabla_x\cdot\big(u^*(x,t)p_t(x)\big)=0.$
Thus the SDE has the same marginal density $p_t$. Setting $\varepsilon(t)=0$ removes the diffusion term and gives the probability flow ODE
\[
\frac{dZ_t}{dt}=u^*(Z_t,t).
\]
On a truncated time interval, suppose that the ODE and SDE are well posed and that their forward equations have unique solutions with the same initial law. The stochastic interpolant, the SDE, and the probability flow ODE then share the marginal densities \(\{p_t\}\) at each time, although their joint distributions over entire paths may differ. For a conditioning variable $w$, the construction is obtained by replacing $p_t(x)$ with the conditional density $p_t(x| w)$. The conditional marginal velocity is $u^*(x,t,w)=\mathbb E[\dot X_t| X_t=x,w].$
The corresponding conditional SDE is
\[
dZ_t=\left(u^*(Z_t,t,w)+\varepsilon(t)\nabla_x\log p_t(Z_t| w)\right)dt+\sqrt{2\varepsilon(t)} dB_t,
\]
and the conditional probability flow ODE is
\[
\frac{dZ_t}{dt}=u^*(Z_t,t,w),~~ Z_t\sim p_t(\cdot| w).
\]

% %%%%%%%%%%%%%%%%%%%%%%%%%%%%%%
\subsection{ReLU deep neural networks}
We consider DNNs with ReLU activations, where ReLU refers to the Rectified Linear Unit. Such a ReLU DNN can be represented as
\begin{equation*}
\begin{aligned}
f_0(x)&=x,\\
f_i(x)&=\sigma\left(\mathbf A_i f_{i-1}(x)+\boldsymbol b_i\right),~~\text{for }i=1,\ldots,\mathcal D-1,\\
f_{\theta}(x)&:=f_{\mathcal D}(x)=\mathbf A_{\mathcal D}f_{\mathcal D-1}(x)+\boldsymbol b_{\mathcal D}.
\end{aligned}
\end{equation*}
Here, the activation function is given by $\sigma(x)=\max\{0,x\}$ and is applied componentwise. The parameter vector $\boldsymbol\theta$ collects all network parameters, while $\mathcal D$ denotes the depth of the network. The sequence of layer dimensions is denoted by $(d_0,d_1,\ldots,d_{\mathcal D})$. In particular, $d_0=d$ gives the input dimension, $d_{\mathcal D}=q$ gives the output dimension, and
$\mathcal W=\max\limits_{1\le i\le\mathcal D}d_i.$
Under this indexing convention, the network has $\mathcal D-1$ hidden layers and $\mathcal D+1$ layers in total when the input layer is included. The parameter set associated with $f_{\theta}$ is therefore written as
\[
\boldsymbol\theta:=\big((\mathbf A_1,\boldsymbol b_1),(\mathbf A_2,\boldsymbol b_2),\ldots,(\mathbf A_{\mathcal D},\boldsymbol b_{\mathcal D})\big).
\]
To describe the componentwise output bound, we write
\[
f_{\theta}(x)=\big(f_{\theta,1}(x),\ldots,f_{\theta,q}(x)\big)^\top,
\]
where $f_{\theta,j}$ denotes the $j$-th output component of $f_\theta$. We assume that
$\max\limits_{j\in[q]}\|f_{\theta,j}\|_\infty\le C_{\mathrm{NN}},$
where $[q]:=\{1,\ldots,q\}$ and
$\|f_{\theta,j}\|_\infty:=\sup_{x\in[0,1]^d}|f_{\theta,j}(x)|.$
Next, we characterize the network complexity through the number and magnitude of its nonzero parameters. Let $\operatorname{vec}(\mathbf A)$ denote the columnwise vectorization of a matrix $\mathbf A$. We define the network size $\mathcal C$ by $\mathcal C=\|\boldsymbol\theta\|_0$, where
\[
\|\boldsymbol\theta\|_0=\sum_{i=1}^{\mathcal D}\left(\|\operatorname{vec}(\mathbf A_i)\|_0+\|\boldsymbol b_i\|_0\right).
\]
The maximum absolute value of all parameters in $\boldsymbol\theta$ is given by
\[
\|\boldsymbol\theta\|_\infty=\max\left\{\max_{1\le i\le\mathcal D}\|\operatorname{vec}(\mathbf A_i)\|_\infty,  \max_{1\le i\le\mathcal D}\|\boldsymbol b_i\|_\infty\right\}.
\]
For convenience, we use $\mathcal B$ to denote the weight bound of the DNN, namely,
$\|\boldsymbol\theta\|_\infty\le\mathcal B.$

\begin{definition}[H{\"o}lder Class]\label{def:holder}
Let $s=r+\tau>0$, where $r\in\mathbb{N}_0$ and $\tau\in(0,1]$. For $d\in\mathbb{N}$, we define the H{\"o}lder class $\mathcal{H}^{s}(\mathbb{R}^d,\Sigma)$ by
\[
\mathcal{H}^{s}(\mathbb{R}^d,\Sigma):=\left\{g:\mathbb{R}^d\to\mathbb{R} :\max_{\|\xi\|_1\le r}\|\partial^\xi g\|_\infty\lesssim \Sigma,  \max_{\|\xi\|_1=r}\sup_{x\ne y}\frac{|\partial^\xi g(x)-\partial^\xi g(y)|}{\|x-y\|^\tau}\lesssim \Sigma\right\}.
\]
Restricting these functions to the domain $[0,1]^d\subseteq\mathbb{R}^d$, we write
\[
\mathcal{H}^{s}:=\left\{g:[0,1]^d\to\mathbb{R} :g\in\mathcal{H}^{s}(\mathbb{R}^d,\Sigma)\right\}.
\]
\end{definition}

\begin{definition}[Total Variation Distance]
	For two probability distributions $\mu$ and $\nu$ on $\Omega \subseteq \mathbb{R}^d$, the total variation distance measures the difference in the probabilities they assign to the same measurable set:
	\[
	\TV(\mu, \nu) = \sup_{\substack{D \subseteq \Omega}} \left| \mu(D) - \nu(D) \right|.
	\]
    Equivalently, writing $\| g \|_{\infty, \Omega} := \sup\limits_{x \in \Omega} |g(x)|$ for $g:\Omega\to\mathbb{R}$, this distance can be expressed by:
	\[
	\TV(\mu, \nu) = \frac{1}{2}\sup_{\substack{\| g \|_{\infty, \Omega} \leq 1}} \left| \int_{\Omega} g(x) \mu(dx) - \int_{\Omega} g(x) \nu(dx) \right|.
	\]
\end{definition}

\subsection{Notations}
For $a,b\ge0$, we write $a\lesssim b$ or $a=\mathcal O(b)$ if $a\le Cb$, and $a\gtrsim b$ or $a=\Omega(b)$ if $a\ge cb$, where $C,c>0$ are constants. We write $a\asymp b$ when both inequalities hold, and use $\widetilde{\mathcal O}$ to suppress logarithmic factors. Let $a\wedge b:=\min\{a,b\}$, and let $\mathbb N$ and $\mathbb N_0$ denote the positive and nonnegative integers, respectively. For vectors, $\|\cdot\|_1$, $\|\cdot\|_2$, and $\|\cdot\|_\infty$ denote the usual norms, and $\|\cdot\|_0$ counts nonzero entries; for matrices, $\|\cdot\|_2$ denotes the spectral norm. For a measurable function $g$, define $\|g\|_{L^2(\mu)}^2:=\int|g|^2 \mathrm d\mu$, $\|g\|_1:=\int|g(x)| \mathrm dx$, and $\|g\|_\infty:=\sup_x|g(x)|$. For a multi-index $\nu\in\mathbb N_0^d$, write $|\nu|:=\sum_{j=1}^d\nu_j$ and let $D^\nu$ denote the corresponding mixed partial derivative. The notation $\nabla_x$ denotes the gradient or Jacobian with respect to $x$, and $\Delta_x$ denotes the Laplacian. Finally, $N(\mathcal F,\varepsilon,\|\cdot\|)$ denotes the covering number of $\mathcal F$ at radius $\varepsilon$ under the indicated norm.

%%%%%%%%%%%%%%%%%%%%%%%%%%%%%%%%%%%%%%%%%%%%%%%%%%%%%%%%%%%%%%%%
%%%%%%%%%%%%%%%%%%%%%%%%%%%%%%%%%%%%%%%%%%%%%%%%%%%%%%%%%%%%%%%%
%%%%%%%%%%%%%%%%%%%%%%%%%%%%%%%%%%%%%%%%%%%%%%%%%%%%%%%%%%%%%%%%
\section{Flow-Based Soft Value Learning}\label{method}\label{sec:integrated-method}
In this section, we propose our framework for estimating the optimal soft action-value function $Q^*$ from offline data by integrating conditional environment distribution learning with empirical Bellman residual minimization. Conditional probability flow ODEs transport Gaussian noise to the conditional reward law $\mathcal R(\cdot|  s,a)$ and transition kernel $\mathcal P(\cdot|  s,a)$. Their velocity fields are estimated by denoising score matching and simulated through EM discretization, yielding the conditional samplers $\widehat{\mathcal R}(\cdot|  s,a)$ and $\widehat{\mathcal P}(\cdot|  s,a)$ from the offline datasets $\mathbb D_r$ and $\mathbb D_x$ (see section~\ref{sec:entropy-mdp}). Samples generated from these learned laws approximate the reward and next state expectations, while importance sampling approximates the log-partition continuation value. These components define a data driven soft optimal Bellman operator $\widehat{\mathcal T}_\alpha$, whose empirical Bellman residual is minimized over a bounded ReLU DNN class on the independent dataset $\mathbb S$ to obtain $\widehat Q$. The resulting procedure forms a complete pipeline from conditional environment distribution learning to soft value function estimation and preserves the sample separation required for the theoretical analysis.
We first define the residual loss under the optimal policy,
\[
\mathcal L_{\pi^*}(\mathcal T_\alpha^*;Q):=\mathbb E_{(S,A)\sim\mu_{\pi^*}}\left|\mathcal T_\alpha^*Q(S,A)-Q(S,A)\right|^2.
\]
If $\mu_{\pi^*}\ll\mu_{\pi_b}$ and the density ratio $\frac{\mathrm d\mu_{\pi^*}}{\mathrm d\mu_{\pi_b}}$ has a finite second moment under $\mu_{\pi_b}$, then by Lemma~\ref{lem:soft-residual-transfer}, we have
$\bigl(\mathcal L_{\pi^*}(\mathcal T_\alpha^*;Q)\bigr)^2\lesssim Q_{\max}^2\mathcal L(\mathcal T_\alpha^*;Q),$
where $\mathcal L(\mathcal T_\alpha^*;Q)$ denotes the population Bellman residual loss for offline observations collected under the behavior policy $\pi_b$, defined by
\begin{equation}\label{eq:loss}
\mathcal L(\mathcal T_\alpha^*;Q):=\mathbb E_{(S,A)\sim\mu}\left|\mathcal T_\alpha^*Q(S,A)-Q(S,A)\right|^2,
\end{equation}
with $\mu=\mu_{\pi_b}$.
Under the H\"older continuity condition on the conditional densities specified in Assumption~\ref{ass:terminal-holder}, we apply Lemmas~\ref{lem:soft-q-holder} and~\ref{lem:soft-residual-transfer} to express our objective as the minimization of the Bellman residual loss under the behavior distribution:
\[
Q^*\in\arg\min_Q\mathcal L(\mathcal T_\alpha^*;Q).
\]
In offline learning, the true operator is unavailable because the conditional reward and transition laws are unknown. The learned conditional samplers above provide its data driven approximation. We present the conditional flow construction for the next-state variable $X=S'$ learned from $\mathbb D_x$. The reward model follows the same construction with $X$ replaced by $R$ and $\mathbb D_x$ by $\mathbb D_r$.

\paragraph{Conditional Flow Learning.}
Let $X=S'\sim p_1(\cdot|  S,A)$ denote the true conditional next-state
variable. With an independent initial noise
$\xi\sim p_0=\mathcal N(0,I_{d_s})$, define the
interpolating path
\[
X_t=tX+\sqrt{1-t}\xi,~~t\in[0,1].
\]
Thus $X_0\sim p_0$ and $X_1\sim p_1(\cdot|  S,A)$. Let
$p_t(\cdot|  S,A)$ denote the conditional density of $X_t$; in particular,
$p_1$ is the transition density introduced in Section~\ref{sec:entropy-mdp}.
The associated conditional probability flow ODE,
$dZ_t=u_x^*(Z_t,t,S,A)dt$, has velocity
\[
u_x^*(x,t,S,A)=\mathbb E[\dot X_t|  X_t=x,S,A]=\frac{x}{t}+\frac{2-t}{2t}\nabla_x\log p_t(x|  S,A).
\]
The score in this expression is unknown, so the velocity is estimated on
$\mathbb D_x$ by denoising score matching. On the truncated
interval $[\tau,1-\tau]$, its population loss is
\[
\mathcal L_x(u)=\frac{1}{1-2\tau}\int_\tau^{1-\tau}\mathbb E_{S,A}\mathbb E_{X_t}\left\|u(X_t,t,S,A)-\frac{X_t}{t}-\frac{2-t}{2t}\nabla_{X_t}\log p_t(X_t|  S,A)\right\|^2dt.
\]
Equivalently, the denoising form uses the target
$X-\xi/(2\sqrt{1-t})$. Given observations $(S_i,A_i,X_i)$ from $\mathbb D_x$,
independently drawn $t_i\sim U[\tau,1-\tau]$, and
Gaussian noises $\xi_i$, we use the empirical objective
\[
\widehat u_x\in\arg\min_{u\in\mathcal U_x}\frac{1}{m}\sum_{i=1}^m\left\|u(t_iX_i+\sqrt{1-t_i}\xi_i,t_i,S_i,A_i)-X_i+\frac{\xi_i}{2\sqrt{1-t_i}}\right\|^2,
\]
where $\{t_i\}_{i=1}^m\sim U[\tau,1-\tau]$ and $\mathcal U_x$ is a class of
ReLU neural networks. Replacing $X_i$ with $R_i$ in this objective and using
$\mathbb D_r$ produces the reward velocity estimator $\widehat u_r$.
Thus, the reward and transition models share the same flow learning
construction but are fitted independently.
To generate samples for a fixed $(s,a)$, we use the EM discretization on the
grid
\[
t_1=\tau,~~t_k=\tau+(k-1)\Delta,~~\Delta=\frac{1-2\tau}{K-1}.
\]
For the transition model, the discretized conditional flow is
\[
\widehat X_{t_{k+1}}=\widehat X_{t_k}+\Delta \widehat u_x(\widehat X_{t_k},t_k,s,a),~~k=1,\ldots,K-1.
\]
The reward trajectories are generated by an analogous update for $\widehat R_t$, with $\widehat u_r$ replacing $\widehat u_x$. The state and reward spaces are $[0,1]^{d_s}$ and $[0,R_{\max}]$, respectively. Upon completing the discretized flow simulation, we clip the generated next-state and reward samples to their respective domains using $\operatorname{CLIP}(z,l,u):=\min\{u,\max\{l,z\}\}$ for $l<u$, applied componentwise when $z$ is a vector. With independent flow inputs at time $\tau$, the clipped terminal outputs yield independent samples $\{\widehat R_i\}_{i=1}^m\sim\widehat{\cR}(\cdot\mid s,a)$ and $\{\widehat S_i'\}_{i=1}^m\sim\widehat{\cP}(\cdot\mid s,a)$.

\paragraph{Operator Construction and Residual Minimization.}
Given $Q\in\mathcal Q$, draw the auxiliary action sample
$\mathbb D_a=\{\widetilde A_j\}_{j=1}^M$ independently from the proposal
density $f$, where $0<c_1\leq f(a)\leq c_2<\infty$ on $\mathcal A$. The
continuation value is approximated by
\[
\widehat V_Q(s)=\alpha\log\bigg[\frac{1}{M}\sum_{j=1}^M\frac{\exp\big(Q(s,\widetilde A_j)/\alpha\big)}{f(\widetilde A_j)}\bigg].
\]
Using the reward and next-state samples generated by the conditional flows,  we can define the empirical optimal Bellman operator
\[
\widehat{\cT}_\alpha Q(s,a)=\frac{1}{m}\sum_{i=1}^m\widehat R_i+\frac{\gamma\alpha}{m}\sum_{i=1}^m\log\bigg[\frac{1}{M}\sum_{j=1}^M\frac{\exp\big(Q(\widehat S_i',\widetilde A_j)/\alpha\big)}{f(\widetilde A_j)}\bigg].
\]
For the subsequent analysis, write $\mathbb D:=(\mathbb D_r,\mathbb D_x,\mathbb D_a)$. Let $\mathbb T$ denote the collection of random time points used to train the reward and transition models, and let $\mathbb Z$ collect the Gaussian noises used in their training and the independent Gaussian inputs used for sample generation. These auxiliary random draws are independent of the offline datasets, and $(\mathbb D,\mathbb T,\mathbb Z)$ is independent of the value learning dataset $\mathbb S$. We regard the generated samples as functions of $(s,a)$ and these fixed random inputs, using the same inputs and action sample for all candidate functions $Q$. Thus, conditional on $(\mathbb D,\mathbb T,\mathbb Z)$, $\widehat{\mathcal T}_\alpha$ is a fixed operator. By Lemma~\ref{lem:T-widehat-gamma}, the learned operator is a $\gamma$-contraction with respect to the supremum norm. 
%%%
We therefore define the corresponding soft Bellman residual loss as
\[
\mathcal L(\widehat{\cT}_\alpha;Q)=\mathbb E_{(S,A)\sim\mu}\left|Q(S,A)-\widehat{\cT}_\alpha Q(S,A)\right|^2.
\]
Using the independent value learning dataset $\mathbb S$, we define the final estimator $\widehat Q$ as the following empirical risk minimizer
\[
\widehat Q\in\arg\min_{Q\in\mathcal Q}\widehat{\mathcal L}(\widehat{\cT}_\alpha;Q):=\frac{1}{n}\sum_{i=1}^n\left|Q(S_i,A_i)-\widehat{\cT}_\alpha Q(S_i,A_i)\right|^2.
\]
As shown in Section~\ref{sec:entropy-mdp}, $Q$ is bounded by $\qm$. Without loss of generality, we restrict $\cQ$ to functions uniformly bounded by $\qm$.
Within the proposed framework, the learned reward and transition models provide an estimate of the optimal Bellman operator. Estimating $Q^\ast$ then reduces to minimizing the soft Bellman residual under the learned soft Bellman operator. The overall structure of the proposed method is presented in Algorithm~\ref{algo:integrated-estimation}.
\begin{algorithm}[H]
\caption{Conditional Flow Models for Offline Maximum Entropy RL}
\label{algo:integrated-estimation}
\begin{algorithmic}[1]
\REQUIRE $\mathbb D_r,\mathbb D_x,\mathbb S$; $K,\tau,m,M,f$.
\ENSURE $\widehat Q$.
\STATE Fit $\widehat u_r$ and $\widehat u_x$ on $\mathbb D_r$ and $\mathbb D_x$, respectively, by denoising score matching.
\STATE Set $\Delta=\frac{1-2\tau}{K-1}$ and $t_k=\tau+(k-1)\Delta$.
\STATE For each $(s,a)$, independently initialize $m$ reward and next-state trajectories from $\mathcal N(0,1-\tau)$ and $\mathcal N(0,(1-\tau)I_{d_s})$, respectively.
\FOR{$k=1,\ldots,K-1$}
\STATE $\widehat R_{t_{k+1}}\leftarrow\widehat R_{t_k}+\Delta\widehat u_r(\widehat R_{t_k},t_k,s,a)$.
\STATE $\widehat S'_{t_{k+1}}\leftarrow\widehat S'_{t_k}+\Delta\widehat u_x(\widehat S'_{t_k},t_k,s,a)$.
\ENDFOR
\STATE Clip terminal outputs to $[0,R_{\max}]$ and $\mathcal S$ to obtain $\{\widehat R_i,\widehat S_i'\}_{i=1}^m$.
\STATE Draw $\mathbb D_a=\{\widetilde A_j\}_{j=1}^M$ from $f$ and construct $\widehat{\cT}_\alpha$ using the generated samples and $\mathbb D_a$.
\STATE Obtain $\widehat Q\in\arg\min\limits_{Q\in\mathcal Q}\widehat{\mathcal L}(\widehat{\cT}_\alpha;Q)$ using $\mathbb S$.
\end{algorithmic}
\end{algorithm}

\section{Theoretical Analysis}\label{sec:theory}
In this section, we establish theoretical guarantees for the estimator $\widehat Q$. Conditional probability flow ODEs estimate the reward law and transition kernel, which determine the learned soft Bellman operator; minimizing the empirical soft Bellman residual associated with this operator then yields $\widehat Q$. The analysis separates the error in learning the soft Bellman operator from the error arising in soft value function learning.

\begin{assumption}\label{ass:conditional-support}
The conditional distribution $p_1(\cdot| s,a)$ is supported on $[0,1]^{d_s}$. 
\end{assumption}

\begin{assumption}\label{ass:score-estimate}
There exists a score estimate $s(x,t,s,a)$ learned from $\bD_x,\bT,\bZ$ such that
\[
\mathbb E_{\bD_x,\bT,\bZ}\mathbb E_{(S,A)}\int_\tau^{1-\tau}\mathbb E_{x_t\sim p_t(\cdot| S,A)}\left\|s(x_t,t,S,A)-\nabla_{x_t}\log p_t(x_t| S,A)\right\|_2^2dt\le \zeta^2 .
\]
\end{assumption}

\begin{assumption}\label{ass:score-regularity}
The score estimate $s(x,t,s,a)$ is $C^2$ with respect to $(x,t)$ for $t\in[\tau,1-\tau]$. There exists $L_\tau\ge 1$ such that, uniformly over $(s,a)$, $\|s(x,t,s,a)\|_2\le L_\tau(1+\|x\|_2),$
and
$\sup\limits_{x\in\mathbb R^{d_s}}\max\limits_{1\le k+|\nu|\le2}\max\limits_{1\le j\le d_s}|\partial_t^k\partial_x^\nu s^{(j)}(x,t,s,a)|\le L_\tau.$ Here $k\in\mathbb N_0$ and $\nu\in\mathbb N_0^{d_s}$.
\end{assumption}

\begin{assumption}\label{ass:terminal-holder}
Let $d=d_s+d_a$ and $w=(s,a)$. For some $\beta>0$ and $\Sigma>0$, the conditional transition density $p_1(x\mid w)$ and the conditional reward density $p_R(r\mid w)$ belong to the H\"older classes $\mathcal H^\beta([0,1]^{d_s+d},\Sigma)$ and $\mathcal H^\beta([0,R_{\max}]\times[0,1]^d,\Sigma)$. 
\end{assumption}

\begin{remark}
Assumption~\ref{ass:conditional-support} imposes a bounded support condition on the conditional target distribution, which is consistent with the normalized state space. In Assumption~\ref{ass:score-estimate}, we assume that the learned score $s(x,t,s,a)$ approximates the conditional score $\nabla_x\log p_t(x| S,A)$ in an integrated $L^2$ sense. This type of score error condition is standard in probability flow ODE analyses \citep{huang2025convergence}. There are several ways to estimate the score function in practice, including score matching, denoising score matching, neural network approximation, and regularized score estimation. These methods have been used to control score approximation and statistical errors in diffusion and probability flow models \citep{oko2023diffusion,cai2025minimax}. Assumption~\ref{ass:score-regularity} is a regularity condition on the learned score, used to control the derivative terms in the Gagliardo--Nirenberg interpolation step.
\end{remark}

We first state separate guarantees for the two sources of error in the proposed estimator. The first result controls the error introduced by replacing the true soft optimal Bellman operator with the operator constructed from the learned reward and transition laws. The second result controls the statistical and neural network approximation errors arising from empirical Bellman residual minimization. For simplicity, in the following Theorems~\ref{thmm:rl-flow-bound}--\ref{thm:main-result}, we write \mbox{$Q_{\max}:=\frac{R_{\max}+\alpha H_{\max}}{1-\gamma}$} (see Section~\ref{sec:entropy-mdp}).

\begin{theorem}\label{thmm:rl-flow-bound}
Suppose that Assumptions~\ref{ass:conditional-support}--\ref{ass:terminal-holder} hold. Assume that $L_\tau=\cO(\tau^{-1})$ and $\Delta=o(\tau^2)$ for both conditional flow estimators. Let the ReLU DNN class $\mathcal Q$ have $\mathcal C=\cO(n^{\frac{d}{d+2\beta}}\log n)$, $\mathcal D=\cO(\log n)$, $\mathcal W=\cO(n^{\frac{d}{d+2\beta}})$, and $\mathcal B=\cO(n^{\frac{d}{d+2\beta}})$. Then,
\begin{align*}
\bE_{\bD,\bT,\bZ}\left\|\cT^*_\alpha\widehat Q-\widehat{\cT}_\alpha\widehat Q\right\|_{L^2(\mu)}^2
&\lesssim Q_{\max}^2\Big(\frac{\zeta}{\tau^4} + \frac{\Delta^2}{\tau^8}\log^2 \tau+\tau^{\beta\wedge1}\Big)+\frac{Q_{\max}^2}{m}n^{\frac{d}{d+2\beta}}\log^2n\log(mn)\\
&~~+\frac{\exp\left(\frac{4Q_{\max}}{\alpha}\right)}{M}n^{\frac{d}{d+2\beta}}\log^2n\log(Mn),\\
\bE_{\bD,\bT,\bZ}\left\|\widehat{\cT}_\alpha Q^*-\cT^*_\alpha Q^*\right\|_{L^2(\mu)}^2
&\lesssim Q_{\max}^2\Big(\frac{\zeta}{\tau^4}
+ \frac{\Delta^2}{\tau^8}\log^2 \tau+\tau^{\beta\wedge1}\Big)+\frac{\exp\left(\frac{Q_{\max}}{\alpha}\right)}{M}+\frac{Q_{\max}^2}{m}.
\end{align*}
\end{theorem}
Theorem~\ref{thmm:rl-flow-bound} shows how conditional distribution estimation and numerical approximation propagate to the learned soft Bellman operator. The next theorem isolates the value learning contribution.

\begin{theorem}\label{thm:rl-error}
Suppose that the conditional reward and transition densities satisfy Assumption~\ref{ass:terminal-holder}, and the class $\mathcal{Q}$ is configured with size $\mathcal{C} = \mathcal{O}(n^{\frac{d}{d + 2\beta}} \log n)$, depth $\mathcal{D} = \mathcal{O}(\log n)$, width $\mathcal{W}=\mathcal{O}(n^{\frac{d}{d + 2\beta}})$, and weight bound $\mathcal{B} = \mathcal{O}(n^{\frac{d}{d + 2\beta}})$. Then,
\begin{equation*}
2\bE_{\bS}\sup_{Q\in\mathcal Q}\big(\cL(\widehat{\cT}_\alpha;Q)-2\widehat{\cL}(\widehat{\cT}_\alpha;Q)\big)+8(1+\gamma)^2\inf_{Q\in\mathcal Q}\|Q-Q^*\|_\infty^2\lesssim \qm^2 n^{-\frac{2\beta}{d+2\beta}}\log^3 n.
\end{equation*}
\end{theorem}
Combining Theorems~\ref{thmm:rl-flow-bound} and~\ref{thm:rl-error} through the error decomposition in Lemma~\ref{lem:error-dec} yields the overall excess Bellman residual guarantee stated in Theorem~\ref{thm:main-result} below.

\begin{theorem}\label{thm:main-result}
Assume the conditions of Theorem~\ref{thmm:rl-flow-bound} and Theorem~\ref{thm:rl-error} hold. Then,
\begin{align*}
\mathbb{E}_{\bS,\bD, \bT,\bZ}\Big[\mathcal L(\mathcal T_\alpha^*;\widehat Q)-\mathcal L(\mathcal T_\alpha^*;Q^*) \Big]&\lesssim \qm^2 n^{-\frac{2\beta}{d+2\beta}}\log^3 n+Q_{\max}^2\Big(\frac{\zeta}{\tau^4}+ \frac{\Delta^2}{\tau^8}\log^2 \tau+\tau^{\beta\wedge1}\Big)\\
& ~~ +\frac{Q_{\max}^2}{m}n^{\frac{d}{d+2\beta}}\log^2n\log(mn)\\
&~~+\frac{\exp\left(\frac{4Q_{\max}}{\alpha}\right)}{M}n^{\frac{d}{d+2\beta}}\log^2n\log(Mn).
\end{align*}
Furthermore, let $m=\Omega(n),~M=\Omega(n)$ and choose $\tau=n^{-\frac{2\beta}{(d+2\beta)(\beta\wedge1)}},~\zeta=n^{-\frac{2\beta(\beta\wedge1)+8\beta}{(d+2\beta)(\beta\wedge1)}},~\Delta=n^{-\frac{\beta(\beta\wedge1)+8\beta}{(d+2\beta)(\beta\wedge1)}}$. Then we have
\[
\mathbb{E}_{\bS,\bD, \bT,\bZ}\Big[\mathcal L(\mathcal T_\alpha^*;\widehat Q)-\mathcal L(\mathcal T_\alpha^*;Q^*) \Big] \lesssim \Big[Q_{\max}^2+\exp\big(\frac{4Q_{\max}}{\alpha}\big)\Big]n^{-\frac{2\beta}{d+2\beta}}\log^3n.
\]
\end{theorem}

\begin{remark}\label{rem:thm-value-est} 
Theorem~\ref{thm:main-result} separates the excess Bellman residual risk into value learning and operator-learning errors. The value learning term $\qm^2 n^{-\frac{2\beta}{d+2\beta}}\log^3 n$ combines the statistical and ReLU approximation errors. The operator learning error consists of the conditional flow error $Q_{\max}^2\big(\frac{\zeta}{\tau^4}+\frac{\Delta^2}{\tau^8}\log^2\tau+\tau^{\beta\wedge1}\big)$, corresponding respectively to score estimation, EM discretization, and terminal smoothing, together with the Monte-Carlo error and the importance sampling error. Ignoring logarithmic factors, the resulting rate is $\mathcal O\big(n^{-\frac{2\beta}{d+2\beta}}\big)$. 

For context, existing entropy regularized RL theory reports rates under rather different settings. \citet{uehara2023offline} study offline minimax soft Q learning and obtain a $\widetilde{\mathcal O}(n^{-1/4})$ statistical error for soft $Q$ estimation and policy regret under realizability and partial coverage. \citet{dai2018sbeed} propose Smoothed Bellman Error Embedding for reinforcement learning with nonlinear function approximation and obtain a statistical Bellman error term $\mathcal O(n^{-1/2})$ for a single $\beta$-mixing off-policy trajectory.  These results are not directly comparable with our excess Bellman residual bound, but they give useful reference points for rates in entropy regularized reinforcement learning.

A key feature of our theoretical analysis is that it does not require Bellman completeness. The approximation error depends only on the fixed target $Q^*$ through $\inf\limits_{Q\in\mathcal Q}\|Q-Q^*\|_\infty^2$, rather than on the Bellman images of all functions in $\mathcal Q$. Consequently, the H\"older regularity of $Q^*$ and the approximation capacity of ReLU DNNs suffice to control this error. This feature follows from estimating the soft Bellman operator before performing Bellman residual minimization and distinguishes our approach from recursive methods that repeatedly project Bellman updates onto a prescribed function class. This approach also differs from existing minimax formulations and methods based on primal and dual optimization for offline reinforcement learning with entropy regularization \citep{uehara2023offline,zhan2022offline}.
\end{remark}

%%%%%%
\section{Numerical Experiments}\label{sec:experiment}
In this section, we consider three groups of offline control experiments. We first study a fully specified simulated MDP with a three-component transition kernel and a two-component reward law, then consider three classical continuous-control tasks, and finally evaluate on four HalfCheetah datasets from D4RL. In every group, the proposed method is compared with offline SAC \citep{haarnoja2018sac}, CQL \citep{kumar2020conservative}, and MOPO \citep{yu2020mopo} using fixed offline data. Evaluation trajectories are never added to the training datasets.

\subsection{Simulated multimodal MDP.}\label{sec:experiment-simulation}
We construct a bounded control problem in which both the transition and reward distributions are multimodal. The state is $s=(x,y,v,c)$, where $x$ is forward position, $y$ is lateral position, $v$ is lateral velocity, and $c$ is an observed context. The domains are
$ \cS=[-2,2]\times[-1,1]\times[-0.45,0.45]\times[-1,1],
    \cA=[-1,1].$
The context determines the target lane,
\begin{equation*}
    y_\star(c)=
    \begin{cases}
        -0.7, & c<-0.25,\\
        0, & |c|\leq0.25,\\
        0.7, & c>0.25.
    \end{cases}
\end{equation*}
The initial coordinates $(x_0,y_0,v_0)$ are independent truncated normal variables with means $(-1.75,0,0)$ and standard deviations $(0.05,0.04,0.025)$. Moreover,
$c_0=\operatorname{clip}_{[-1,1]}(0.8Z+\varepsilon_c)$, where $Z$ is uniform on $\{-1,1\}$ and $\varepsilon_c\sim\mathcal N(0,0.02^2)$. Thus, each trajectory starts near the center lane while its observed context usually selects an outer lane. The horizon is 180 and $\gamma=0.99$. The simulator uses the reference command
\begin{equation*}
    a_{\mathrm{ref}}(s)=\operatorname{clip}_{[-1,1]}
       \bigl(y_\star(c)+0.90(y_\star(c)-y)-0.45v\bigr).
\end{equation*}
This command defines the dynamics and reward but is not included as a label in the offline data. Before boundary adjustment, the central next-state location $\bar\mu(s,a)=(\bar x',\bar y',\bar v',\bar c')$ is
\begin{align*}
    &\bar x'=x+0.021+0.007\exp \left[-\frac{(a-a_{\mathrm{ref}}(s))^2}{2(0.30)^2}\right],~~
    \bar y'=0.78y+0.22a+0.04v,\\
   & \bar v'=0.45v+0.16(a-y_\star(c))+0.12(y_\star(c)-y),~~
    \bar c'=c.
\end{align*}
Let
$ d=(0,0.025,0.015,0),
    \sigma=(0.0025,0.006,0.005,0.001),$
and let $\ell,u$ be the coordinatewise bounds of $\cS$. For $j\in\{-1,0,1\}$, define
$ \mu_j(s,a)=\operatorname{clip}_{[\ell+4\sigma, u-4\sigma]}
       \bigl(\bar\mu(s,a)+jd\bigr).$
The transition kernel is
\begin{align*}
    P(\cdot|  s,a)
    &=0.25\operatorname{TN}_{\cS}\bigl(\mu_{-1}(s,a),\operatorname{diag}(\sigma^2)\bigr)
    +0.50\operatorname{TN}_{\cS}\bigl(\mu_{0}(s,a),\operatorname{diag}(\sigma^2)\bigr)\\
    &~~ +0.25\operatorname{TN}_{\cS}\bigl(\mu_{1}(s,a),\operatorname{diag}(\sigma^2)\bigr),
\end{align*}
where $\operatorname{TN}_{\cS}$ is a multivariate normal distribution truncated to $\cS$. The component is resampled independently at each transition; rejection sampling keeps every coordinate within its stated bound. For the reward law, define
\begin{align*}
    g_a(s,a)=\exp \left[-\frac{(a-a_{\mathrm{ref}}(s))^2}{2(0.22)^2}\right],
    g_y(s)=\exp \left[-\frac{(y-y_\star(c))^2}{2(0.18)^2}\right],
    g_v(s)=\exp \left[-\frac{v^2}{2(0.10)^2}\right],
\end{align*}
and $b(s,a)=-5+8g_a(s,a)+5g_y(s)+2g_v(s)
             +g_a(s,a)g_y(s)-0.08a^2.$
Conditional on $(s,a)$, the observed reward follows the symmetric mixture
\begin{equation*}
    R| (s,a)\sim
    \tfrac12\operatorname{TN}_{[-20,20]}\bigl(b(s,a)-6.5,0.035^2\bigr)
    +\tfrac12\operatorname{TN}_{[-20,20]}\bigl(b(s,a)+6.5,0.035^2\bigr).
\end{equation*}
The offline data are generated by a behavior policy that samples $a$ uniformly from $[-1,1]$. For each record, an independent initial state is advanced for $L\sim\operatorname{Unif}\{0,\ldots,179\}$ behavior steps, after which one tuple $(s,a,r,s')$ is retained. Independent random streams produce $|\mathbb D_r|=100\mathrm{K}$ reward samples, $|\mathbb D_x|=100\mathrm{K}$ transition samples, and $|\mathbb S|=790\mathrm{K}$ samples for value learning. 
After training, we evaluate each deterministic policy over 100 independent trajectories in the stochastic environment. Performance is measured by the discounted return on the original reward scale
\begin{equation*}
    J_{180}(\pi)=\mathbb E_\pi \left[\sum_{t=0}^{179}0.99^tR_t\right].
\end{equation*}
Table~\ref{tab:simulation-final} reports the final discounted returns as the mean $\pm$ standard deviation over 100 evaluation trajectories. The proposed method achieves the highest mean discounted return of 861.69, exceeding SAC, CQL, and MOPO by 328.21, 357.38, and 344.06, respectively. Its standard deviation of 40.37 is also lower than those of SAC (282.65), CQL (135.00), and MOPO (303.31), indicating that the improvement in mean return is accompanied by more consistent performance across evaluation trajectories.
\begin{table}[H]
    \centering
    \small
    \caption{Experimental Results on the simulated MDP.}
    \label{tab:simulation-final}
    \begin{tabular}{cc}
        \toprule
        Method & Return \\
        \midrule
        SAC  & 533.48 $\pm$ 282.65 \\
        CQL  & 504.31 $\pm$ 135.00 \\
        MOPO & 517.63 $\pm$ 303.31 \\
        Ours & \textbf{861.69 $\pm$ 40.37} \\
        \bottomrule
    \end{tabular}
\end{table}

\subsection{Classical continuous-control tasks.}\label{sec:experiment-classical}
In this section, we compare the methods on three tasks: standard Pendulum, a stochastic inverted pendulum, and a stochastic five-mode MountainCar. Ordinary Pendulum follows the deterministic Pendulum-v1 dynamics with state $(\theta,\dot\theta)$, bounded torque, a 200-step horizon, and the standard quadratic control cost. The stochastic version adds an actuator disturbance in $\{-0.45,0,0.45\}$ with probabilities $(0.20,0.60,0.20)$ and Gaussian acceleration noise with standard deviation $0.08$. MountainCar uses position and velocity as its state, terminates upon reaching the goal or after 999 steps, and adds one of five equally likely wind accelerations $(-0.0008,-0.0004,0,0.0004,0.0008)$ at every transition. For Pendulum and stochastic inverted pendulum, the behavior policy chooses a uniform random torque with probability $0.80$ and a noisy upright PD action otherwise. Each retained transition follows $L\sim\operatorname{Unif}\{0,\ldots,79\}$ behavior steps from an independent reset. For MountainCar, the behavior policy mixes uniform actions with a noisy momentum-building controller using probabilities $0.55$ and $0.45$, and $L\sim\operatorname{Unif}\{0,\ldots,399\}$ with terminal resets. The reported score is the raw undiscounted episode return
\begin{equation*}
    G_T(\pi)=\mathbb E_\pi \left[\sum_{t=0}^{T-1}R_t\right].
\end{equation*}
The horizon is set to $T=200$ for the two pendulum tasks, while MountainCar episodes terminate upon reaching the goal or at $T=999$. Table~\ref{tab:classical} reports the mean returns obtained from environment evaluations after training for each task, with higher values indicating better performance.
We train the models for 500K updates on each task, and then evaluate the trained models in their respective environments.
The dataset sizes, $\gamma$, $\alpha$, and the remaining settings of Ours are the same as in the simulated MDP experiments.
\begin{table}[H]
    \centering
    \caption{Experimental Results on classical control tasks.}
    \label{tab:classical}
    \small
    \begin{tabular}{ccccc}
        \toprule
        Task & SAC & CQL & MOPO & Ours \\
        \midrule
        Ordinary Pendulum
        & $-102.761$ & $-114.905$ & $-104.206$ & $\mathbf{-101.176}$ \\
        Stochastic inverted pendulum
        & $-97.397$ & $-101.451$ & $-97.132$ & $\mathbf{-95.289}$ \\
        Stochastic MountainCar
        & $96.025$ & $94.452$ & $ 95.919$ & $\mathbf{96.036}$ \\
        \bottomrule
    \end{tabular}
\end{table}
In  each of the three tasks, the proposed method achieves the highest return: $-101.176$ on Pendulum, $-95.289$ on stochastic inverted pendulum, and $96.036$ on MountainCar. The corresponding improvements over SAC are $1.585$, $2.108$, and $0.011$, respectively, indicating that the MountainCar results are effectively tied. These results suggest that the proposed method offers a performance advantage in the evaluated environments.

\subsection{D4RL HalfCheetah.}\label{sec:experiment-d4rl}
Finally, we evaluate the proposed method on HalfCheetah from the D4RL benchmark \citep{fu2020d4rl}, using fixed offline datasets without additional data collection during optimization. HalfCheetah is a continuous control task in which a planar articulated agent applies bounded joint torques to move forward. The task is sensitive to the learned action distribution, as locally supported actions may eventually lead to states poorly represented in the offline data. We consider four datasets that differ in behavior quality and state-action coverage. Random contains transitions from a random policy, providing broad coverage but low returns, while Medium contains data from a partially trained policy with moderate performance. Medium-Replay comprises the replay buffer accumulated during training of the medium policy, combining behavior from early and later training stages. Full-Replay spans a longer period of policy improvement and provides broader replay coverage.

All methods are trained for 1M gradient updates with the temperature parameter fixed at $\alpha=0.2$. The final results are obtained from the models after 1M updates. Each final policy is evaluated for 100 episodes, and Table~\ref{tab:final-performance} reports the mean and standard deviation of the resulting episode returns. The proposed method uses a conditional Gaussian proposal distribution parameterized by the current state, followed by a hyperbolic tangent transformation. A latent action $z$ is sampled from a diagonal Gaussian distribution, and the bounded action $a$ is obtained as
\begin{equation*}
    z \sim \mathcal{N} \left(
        \mu_{\theta}(s),
        \operatorname{diag} \left(\sigma_{\theta}^{2}(s)\right)
    \right),
    ~a = \tanh(z).
\end{equation*}
The mean depends on the current state and determines where the proposal probability is concentrated. The scale parameter controls the dispersion of candidate actions around that region. The transformation ensures that each sampled action satisfies the bounds of the HalfCheetah action space. We report performance using the D4RL normalized score. Given an average undiscounted return $R$, the normalized score is defined as
\begin{equation}
    \operatorname{Score}_{\mathrm{D4RL}}(R)
    =
    100\times 
    \frac{R-R_{\mathrm{random}}}
         {R_{\mathrm{expert}}-R_{\mathrm{random}}},
\label{eq:d4rl-score}
\end{equation}
where $R_{\mathrm{random}}$ and $R_{\mathrm{expert}}$ denote the random policy and expert policy reference returns provided by D4RL for HalfCheetah. This transformation assigns a score of zero to the random policy reference and a score of 100 to the expert policy reference. All methods are normalized using the same reference returns.
\begin{table}[H]
    \centering
    \caption{Experimental Results on D4RL HalfCheetah}
    \label{tab:final-performance}
    \small
    \begin{tabular}{ccccc}
        \toprule
        Dataset & SAC & CQL & MOPO & Ours \\
        \midrule
        Random
          & \textbf{28.31 $\pm$ 0.26}
          & 19.11 $\pm$ 0.08
          & 26.31 $\pm$ 0.27
          & 24.43 $\pm$ 0.15 \\
        Medium
          & 20.04 $\pm$ 17.63
          & 49.42 $\pm$ 0.79
          & 44.44 $\pm$ 11.57
          & \textbf{64.03 $\pm$ 10.12} \\
        Medium-Replay
          & 33.03 $\pm$ 16.75
          & 47.02 $\pm$ 0.79
          & \textbf{50.87 $\pm$ 19.95}
          & 48.88 $\pm$ 0.78 \\
        Full-Replay
          & \textbf{78.53 $\pm$ 11.73}
          & 76.63 $\pm$ 0.73
          & 67.68 $\pm$ 19.72
          & 71.94 $\pm$ 0.69 \\
        \midrule
        Mean
          & 39.98
          & 48.05
          & 47.32
          & \textbf{52.32} \\
        \bottomrule
    \end{tabular}%
\end{table}
The proposed method achieves an average normalized score of 52.32 across the four datasets, exceeding CQL, SAC, and MOPO by 4.27, 12.34, and 5.00 points, respectively. The largest gain occurs on Medium, where its score of 64.03 exceeds the strongest baseline, CQL, by 14.61 points. On Medium-Replay, it obtains $48.88\pm0.78$, outperforming CQL ($47.02\pm0.79$) and SAC ($33.03\pm16.75$), while remaining below MOPO ($50.87\pm19.95$) with substantially lower evaluation variability. These results suggest that the conditional Gaussian proposal, which concentrates sampling near actions associated with the current state while retaining local stochasticity, is effective under restricted or heterogeneous data coverage. On Random, the proposed method obtains $24.43\pm0.15$, exceeding CQL by 5.32 points but trailing SAC by 3.88 points. On Full-Replay, it achieves $71.94\pm0.69$, below SAC ($78.53\pm11.73$) and CQL ($76.63\pm0.73$). Its advantages are therefore most evident on Medium and Medium-Replay, where useful behavior is available but data coverage remains restricted or heterogeneous.

\section{Conclusion}\label{sec:conclusion}

This work studies the estimation of the optimal soft action value function \(Q^*\) from offline data in maximum entropy reinforcement learning. We use conditional probability flow ODEs to learn the reward and transition laws and construct an approximation of the soft optimal Bellman operator. The value function is then fitted by minimizing the empirical residual of this operator. Learning the environment model before fitting the value function allows us to distinguish the errors introduced at each stage and study their effect on the final residual risk. The analysis relies on approximating \(Q^*\) itself, without imposing Bellman completeness on the neural network class.

For the value learning stage, the statistical error and the ReLU approximation error together contribute \(\mathcal O\bigl(n^{-\frac{2\beta}{d+2\beta}}\bigr)\) to the excess Bellman residual risk, up to logarithmic factors. The remaining error comes from approximating the soft Bellman operator. Its conditional flow component is bounded by \(\mathcal O\bigl(\frac{\zeta}{\tau^4}+\frac{\Delta^2}{\tau^8}+\tau^{\beta\wedge1}\bigr)\), with the three terms arising from score estimation, numerical discretization, and terminal smoothing. Monte-Carlo averaging over generated rewards and next states adds a contribution of \(\mathcal O\bigl(\frac{n^{\frac{d}{d+2\beta}}}{m}\bigr)\), while importance sampling for the soft state value introduces \(\mathcal O\bigl(\frac{n^{\frac{d}{d+2\beta}}}{M}\bigr)\). Logarithmic factors are omitted from these expressions.

With the assumptions and parameter choices in the main theorem, the operator learning terms are controlled at the same order as the value learning term. The resulting excess Bellman residual risk therefore has the nonparametric rate \(\mathcal O\bigl(n^{-\frac{2\beta}{d+2\beta}}\bigr)\), up to logarithmic factors, without requiring Bellman completeness. The numerical experiments complement these guarantees and illustrate the effectiveness of the method on the tasks considered. Building on these results, future work could study offline policy optimization and examine how the approach scales to higher dimensional continuous control. Extending the framework to partially observed environments and larger world models would also broaden its range of applications.

%%%%%%%%%%%%%%%%%%%%%%%%%%%%%%%%%%%%%%%%
 
\begin{appendix}
\section*{Appendix} 
The appendices follow the structure of the theoretical analysis. Appendices~\ref{sec:soft-bellman} and~\ref{sec:value learning} prove the operator learning and value learning theorems, respectively. Building on these results, Appendix~\ref{sec:main-results} characterizes the population residual objective, establishes the error decomposition lemma, and proves the main theorem. Appendix~\ref{sec:apd-auxiliary-results} collects the auxiliary lemmas used in these proofs.

%%%%%%%%%%%%%%%%附录目录%%%%%%%%%%%%%%%%
\startcontents[appendix]
\printcontents[appendix]{l}{1}{\subsection*{Table of Contents}\setcounter{tocdepth}{2}}

\section{Proof Outline}\label{sec:proof-sketch}

The preceding section states the soft operator learning, value learning, and overall guarantees. We now explain how the two intermediate bounds are obtained and how they combine to prove Theorem~\ref{thm:main-result}. The argument begins with the following decomposition of the excess Bellman residual risk; its full proof is provided in Appendix~\ref{sec:main-results}.

\begin{lemma}[Error decomposition]
\label{lem:error-dec}
Conditional on $\bD,\bT,\bZ$, the estimator $\widehat Q$ satisfies
\[
\begin{aligned}
\bE_{\bS}\left[\mathcal L(\mathcal T_\alpha^*;\widehat Q)-\mathcal L(\mathcal T_\alpha^*;Q^*)\right]
&\le \underbrace{2\bE_{\bS}\left[\|\cT^*_\alpha\widehat Q-\widehat{\cT}_\alpha\widehat Q\|_{L^2(\mu)}^2\right]}_{\mathrm{Part}~\mathbf{I}}+\underbrace{8\|\widehat{\cT}_\alpha Q^*-\cT^*_\alpha Q^*\|_{L^2(\mu)}^2}_{\mathrm{Part}~\mathbf{II}}\\
&~ +\underbrace{2\bE_{\bS}\sup_{Q\in\mathcal Q}\Big(\cL(\widehat{\cT}_\alpha;Q)-2\widehat{\cL}(\widehat{\cT}_\alpha;Q)\Big)+8(1+\gamma)^2\inf_{Q\in\mathcal Q}\|Q-Q^*\|_\infty^2}_{\mathrm{Part}~\mathbf{III}}.
\end{aligned}
\]
\end{lemma}
Parts~\textbf{I} and~\textbf{II} evaluate the discrepancy between the true and learned Bellman operators at $\widehat Q$ and $Q^*$, respectively. They quantify how errors in estimating the conditional reward law and transition kernel propagate to the Bellman operator. Part~\textbf{III} contains the statistical error and neural network approximation error of value learning. We first derive the operator learning bound in Theorem~\ref{thmm:rl-flow-bound}, then establish the value learning bound in Theorem~\ref{thm:rl-error}, and finally combine them through Lemma~\ref{lem:error-dec} to obtain Theorem~\ref{thm:main-result}.

\subsection{Error Bounds for Parts I and II}\label{sec:eb1}
In this subsection, we derive upper bounds for Parts~$\mathbf{I}$ and~$\mathbf{II}$ in Lemma~\ref{lem:error-dec}. These terms quantify the error caused by replacing the true soft optimal Bellman operator with its learned counterpart. As shown in Appendix~\ref{sec:B.5}, their bounds depend on the estimation errors of the conditional reward and transition distributions, together with the Monte-Carlo approximation errors in the construction of $\widehat{\cT}_\alpha$. We begin by establishing a total variation bound for the conditional probability flow estimator.

Since the true conditional distributions are supported on their prescribed domains, coordinatewise clipping leaves them unchanged. By the data-processing inequality for total variation, clipping the generated outputs therefore cannot increase their TV errors relative to these target distributions. Hence, it suffices to analyze the flow outputs before clipping. The total variation error can be decomposed as
\begin{equation}\label{eq:TV-decomposition}
\begin{aligned}
\TV^2\Big(\widehat p_{t_K}(\cdot|  S,A),p_1(\cdot|  S,A)\Big)
&\lesssim \TV^2\Big(\widehat p_{t_K}(\cdot|  S,A),p_{1-\tau}(\cdot|  S,A)\Big)\\
&~~+\TV^2\Big(p_{1-\tau}(\cdot|  S,A),p_1(\cdot|  S,A)\Big).
\end{aligned}
\end{equation}
Here, $p_{1-\tau}(\cdot|  S,A)$ is the marginal density generated by the exact probability flow at time $1-\tau$, while $\widehat p_{t_K}(\cdot|  S,A)$ is the density generated by the learned flow under the EM scheme before clipping, with $t_K=1-\tau$. The first term in \eqref{eq:TV-decomposition} captures the combined effects of Gaussian initialization, score estimation, and numerical discretization. The second term is the terminal smoothing bias resulting from stopping the flow at time $1-\tau$. We bound these two terms separately below.

\paragraph{Bound $\TV^2\big(\widehat p_{t_K}(\cdot|  S,A),p_{1-\tau}(\cdot|  S,A)\big)$} \label{sec:TV-bound1}

In this part, we bound the first term in \eqref{eq:TV-decomposition}. This term compares the EM discretized conditional flow induced by the learned score estimate with the exact conditional probability flow. For each fixed $(s,a)$, the exact density $p_t(\cdot|  s,a)$ solves the transport equation with velocity field $u^*(\cdot,t,s,a)$, while the EM interpolated density $\widehat p_t(\cdot|  s,a)$ solves a transport equation with the frozen velocity field $\widehat u(\Phi_{k,t}^{-1}(\cdot),t_k,s,a)$ on each interval $[t_k,t_{k+1}]$.
In addition to the initial density difference caused by Gaussian initialization, the transport equations introduce two error terms: the score-estimation error
$\widehat u(x,t,s,a)-u^*(x,t,s,a)=\frac{2-t}{2t}\big(s(x,t,s,a)-\nabla_x\log p_t(x|  s,a)\big)$,
and the EM discretization error
$\widehat u(\Phi_{k,t}^{-1}(x),t_k,s,a)-\widehat u(x,t,s,a)$.
Following the $L^1$ stability strategy for probability flow ODEs in \citep{huang2025convergence}, we control the density difference by the transport equation stability estimate. The score estimation part is bounded by the Gagliardo--Nirenberg interpolation inequality, while the discretization part is controlled by the regularity of the learned velocity field and the step size $\Delta$. This yields the following conditional learned EM flow error bound.

\begin{lemma}\label{lem:tv-score}
Suppose Assumptions~\ref{ass:conditional-support}, \ref{ass:score-estimate}, and~\ref{ass:score-regularity} hold. 
%Initialize the learned EM flow at time $\tau$ with the density $q_\tau$ of $\mathcal N(0,(1-\tau)I_{d_s})$ as in Algorithm~\ref{algo:integrated-estimation}. 
Assume that $L_\tau=\cO(\tau^{-1})$ and $\Delta=o(\tau^2)$ as $\tau$ tends to zero. Then
\[
\mathbb E_{\bD_x,\bT,\bZ}\mathbb E_{(S,A)}\TV^2\Big(\widehat p_{t_K}(\cdot|  S,A),p_{1-\tau}(\cdot|  S,A)\Big) 
\lesssim
%\frac{d_s\tau^2}{1-\tau}+\frac{\zeta}{\tau^4} 
\frac{\tau^2}{1-\tau}+\frac{\zeta}{\tau^4} 
+ \frac{\Delta^2}{\tau^8}\log^2 \tau.
\]
\end{lemma}

\begin{remark}\label{rem:tv-score-structure}
The upper bound in Lemma~\ref{lem:tv-score} separates Gaussian initialization error, score estimation error, and time discretization error. The initialization term is 
% $\mathcal O\big(d_s\tau^2/(1-\tau)\big)$,
$\mathcal O\big(\tau^2/(1-\tau)\big)$,
as established directly in the proof. The term $\mathcal{O}\big(\frac{\zeta}{\tau^4}\big)$ accounts for the error from approximating the exact velocity field with the learned score function, while $\mathcal{O}\big(\frac{\Delta^2}{\tau^8}\log^2\tau\big)$ captures the numerical error induced by the discrete EM scheme. Similar score and discretization error decompositions appear in probability flow ODE analyses \citep[e.g.,][]{huang2025convergence,cai2025minimax,gao2025convergence}.
\end{remark}

\paragraph{Bound $\TV^2 \big(p_{1-\tau}(\cdot|S,A),p_{1}(\cdot|S,A)\big)$}\label{sec:TV-bound2}

We now bound the second term in the decomposition, which is the terminal time truncation bias caused by stopping the forward interpolation at time $1-\tau$ instead of the terminal time $1$. For each fixed $(S,A)$, the random variable $X_{1-\tau}=(1-\tau)X+\sqrt{\tau}\xi$ is a Gaussian-smoothed and slightly contracted version of $X\sim p_1(\cdot|  S,A)$. Hence the discrepancy between $p_{1-\tau}(\cdot|  S,A)$ and $p_1(\cdot|  S,A)$ is controlled by the H\"older smoothness of the terminal density on $[0,1]^{d_s}$ and by the boundary leakage induced by the Gaussian perturbation. This gives the following terminal smoothing bias bound.

\begin{lemma}\label{lem:tv-time}
Under Assumptions \ref{ass:conditional-support} and \ref{ass:terminal-holder},
\[
\mathbb E_{(S,A)}\TV^2\Big(p_{1-\tau}(\cdot|  S,A),p_1(\cdot|  S,A)\Big)\lesssim \tau^{\beta\wedge1}.
\]
\end{lemma}
The bounds in Lemmas \ref{lem:tv-score}--\ref{lem:tv-time} together imply the following error bound for transition kernel estimation.
\begin{lemma} \label{lem:TV-flow-x}
Suppose that the conditions of 
Lemmas~\ref{lem:tv-score}--\ref{lem:tv-time} 
 are satisfied. Then,
\begin{equation*}
	\mathbb{E}_{\bD_x,\bT,\bZ}\bE_{(S,A)}\Big[\TV^2 \Big(\widehat{\mathcal{P}}(\cdot|S,A),\mathcal{P}(\cdot|S,A)\Big)\Big]\lesssim \frac{\zeta}{\tau^4} + \frac{\Delta^2}{\tau^8}\log^2 \tau+\tau^{\beta\wedge1}.
\end{equation*}
\end{lemma}
Similarly, we can also derive an error bound for reward law estimation in the following corollary.
\begin{corollary}\label{cor:TV-flow-r}
Under the    
conditions of Lemma~\ref{lem:TV-flow-x} with the same H\"older index $\beta$, 
we have
\begin{equation*}
\mathbb{E}_{\bD_r,\bT,\bZ} \bE_{(S,A)}\Big[\TV^2 \Big(\widehat{\mathcal{R}}(\cdot|S,A),\mathcal{R}(\cdot|S,A)\Big)\Big]\lesssim \frac{\zeta}{\tau^4} + \frac{\Delta^2}{\tau^8}\log^2 \tau+\tau^{\beta\wedge1}.
\end{equation*}
\end{corollary}

Combining Lemma~\ref{lem:error-dec}, Lemma~\ref{lem:TV-flow-x}, and Corollary~\ref{cor:TV-flow-r} establishes the two operator-learning inequalities stated in Theorem~\ref{thmm:rl-flow-bound}.

\begin{remark}\label{rem:operator-tv-structure}
The error bounds in Theorem~\ref{thmm:rl-flow-bound} separate into three main parts: the score estimation error of order $\mathcal O\big(\frac{\zeta}{\tau^4}\big)$, the EM time discretization error of order $\mathcal O\big(\frac{\Delta^2}{\tau^8}\log^2\tau\big)$, and the terminal smoothing bias of order $\mathcal O(\tau^{\beta\wedge1})$. These three terms represent the total variation error of the learned reward and transition models, while the remaining terms in the theorem come from Monte-Carlo sampling and importance sampling in the empirical Bellman operator.
\end{remark}

\subsection{Error Bound for Part III}\label{sec:eb2}
In this section, we analyze Part~$\mathbf{III}$ of the error decomposition in Lemma \ref{lem:error-dec}, which captures the error arising from value learning. It consists of the statistical error
$\sup\limits_{Q\in\mathcal Q}\big(\mathcal L(\widehat{\mathcal T}_\alpha;Q)-2\widehat{\mathcal L}(\widehat{\mathcal T}_\alpha;Q)\big)$
and the neural network approximation error
$\inf\limits_{Q\in\mathcal Q}\|Q-Q^*\|_\infty^2.$
We derive bounds for these two terms separately and then combine them to obtain an upper bound on the Bellman residual risk error.

\paragraph{Bound statistical error $\sup\limits_{Q\in\mathcal Q}\big(\mathcal{L}({\widehat{\mathcal{T}}}_{\alpha};Q)-2\widehat{\mathcal{L}}({\widehat{\mathcal{T}}}_{\alpha};Q)\big)$.} 
To establish the desired bound, we follow the proof strategy of
\citet[Lemma~15]{chen2023score}. Specifically, the infinite function class
$\mathcal Q$ is first approximated by a finite $\delta$-net, thereby reducing
the uniform concentration problem to one over a finite collection of neural
networks. The cardinality of this $\delta$-net is controlled using the
covering number estimate in \citet[Lemma~20]{feng2024deep}. Substituting the
prescribed network size, depth, width, and parameter bound into the resulting
entropy estimate yields the following bound.

\begin{lemma}
\label{lem:term-2-statistical}
Assume that the class $\mathcal Q$ of ReLU DNNs has size $\mathcal C=\mathcal O(n^{\frac{d}{2\beta+d}}\log n)$, depth $\mathcal D=\mathcal O(\log n)$, width $\mathcal W=\mathcal O(n^{\frac{d}{2\beta+d}})$, and weight bound $\mathcal B=\mathcal O(n^{\frac{d}{2\beta+d}})$. Then
\[
\mathbb E_{\mathbb S}\sup_{Q\in\mathcal Q}\bigl(\mathcal L(\widehat{\mathcal T}_\alpha;Q)-2\widehat{\mathcal L}(\widehat{\mathcal T}_\alpha;Q)\bigr)\lesssim Q_{\max}^2n^{-\frac{2\beta}{2\beta+d}}\log^3n.
\]
\end{lemma}

\begin{remark}\label{rem:statistical-error}
To derive the statistical error bound in
Lemma~\ref{lem:term-2-statistical}, we assume that the batch data are i.i.d.,
following the standard setting in the statistical analysis of deep
reinforcement learning
\citep{jiang2016doubly,fan2020theoretical,uehara2022review}.
Our proof applies a concentration inequality to a finite $\delta$-net of the
ReLU DNN class $\mathcal Q$ and then extends the resulting bound to the full
class through a covering-number argument. Under the network architecture and
parameter choices specified in Lemma~\ref{lem:term-2-statistical}, this gives
the rate
\(
  \mathcal O\bigl(Q_{\max}^2n^{-\frac{2\beta}{2\beta+d}}\log^3n\bigr)
\).
The proposal and temperature parameters are fixed with respect to $n$, and their dependence is absorbed into the implicit constant.
Moreover, the i.i.d. assumption can be extended to dependent data by accounting
for the sequential structure of MDP trajectories. A standard approach is to
impose suitable mixing conditions and apply concentration results for
dependent samples to establish the corresponding statistical error bound, as
in \citet{feng2023overparameterized,jiao2025deep}.
\end{remark}

\paragraph{Bound approximation error $\inf\limits_{Q\in\mathcal Q}\|Q-Q^*\|_\infty^2$.}
For the soft optimal Bellman operator $\cT_\alpha^*$, approximate Bellman completeness typically requires the image $\cT_\alpha^*\mathcal Q$ to be uniformly well-approximated by $\mathcal Q$, as quantified by $\sup\limits_{Q\in\mathcal Q}\inf\limits_{\widetilde Q\in\mathcal Q}\|\widetilde Q-\cT_\alpha^*Q\|_\infty$. While standard in iterative value-based methods that repeatedly project Bellman updates onto $\mathcal Q$, this closure condition is not intrinsic to maximum entropy reinforcement learning. Indeed, recent works on entropy-regularized offline RL establish theoretical guarantees by replacing Bellman completeness with realizability conditions tailored to their specific estimators. For instance, \citet{uehara2023offline} impose realizability on both $Q^*$ and an auxiliary saddle-point function, whereas \citet{zhan2022offline} analyze a regularized primal-dual formulation under realizability and single-policy concentrability.

Our analysis takes a more direct route. Because the proposed framework explicitly estimates the reward distribution and transition kernel before minimizing the empirical soft Bellman residual, the error decomposition in Lemma~\ref{lem:error-dec} reduces the approximation overhead to $\inf\limits_{Q\in\mathcal Q}\|Q-Q^*\|_\infty^2$, completely bypassing uniform closure conditions over $\cT_\alpha^*\mathcal Q$. Consequently, it suffices to approximate the single target function $Q^*$, whose smoothness is guaranteed by the following result.

\begin{lemma}\label{lem:soft-q-holder}
Suppose that the conditional reward and transition densities satisfy Assumption~\ref{ass:terminal-holder}. Then the optimal soft action-value function satisfies $Q^*\in\cH^\beta$.
\end{lemma}
With the H\"older regularity from Lemma~\ref{lem:soft-q-holder}, we can directly invoke the deep ReLU network approximation theory of \citet{jiao2025deep} to bound $\inf\limits_{Q\in\mathcal Q}\|Q-Q^*\|_\infty^2$ (see Appendix~\ref{sec:value learning} for the detailed derivation). Combining this approximation rate with the preceding statistical error bound yields Theorem~\ref{thm:rl-error}.

\section{Proof of Theorem~\ref{thmm:rl-flow-bound}}
\label{sec:soft-bellman}
We first address the two operator learning terms in Lemma~\ref{lem:error-dec}. The argument controls the discrepancy between the learned and true conditional reward and transition laws, and then propagates these distributional errors, together with sampling errors, to the soft Bellman operator. This establishes the operator bound stated in Theorem~\ref{thmm:rl-flow-bound}.

\subsection{Proof of Lemma \ref{lem:tv-score}}

\begin{proof}
 By the definition of $\widehat u$, Assumption~\ref{ass:score-regularity}, and $t\in[\tau,1-\tau]$,
\[
\|\widehat u(x,t,s,a)\|_2=\left\|\frac{x}{t}+\frac{2-t}{2t}s(x,t,s,a)\right\|_2\le \frac{1}{\tau}\|x\|_2+\frac{L_\tau}{\tau}(1+\|x\|_2)\lesssim \frac{1+\|x\|_2}{\tau^2}.
\]
Moreover,
\[
\nabla_x\widehat u(x,t_k,s,a)=\frac{1}{t_k}I_{d_s}+\frac{2-t_k}{2t_k}\nabla_xs(x,t_k,s,a),
\]
and hence
\[
\|\nabla_x\widehat u(x,t_k,s,a)\|_2\le \frac{1}{t_k}+\frac{2-t_k}{2t_k}\|\nabla_xs(x,t_k,s,a)\|_2\le \frac{C(1+L_\tau)}{\tau}\lesssim \frac{1}{\tau^2},
\]
where $C>0$ is a universal constant. For $k=1,\ldots,K-1$ and $t\in[t_k,t_{k+1}]$, define
\begin{equation}\label{eq:tv-fun}
\Phi_{k,t}(x)=x+(t-t_k)\widehat u(x,t_k,s,a).
\end{equation}
Then, for any $x,y\in\mathbb R^{d_s}$,
\[
\begin{aligned}
\|\Phi_{k,t}(x)-\Phi_{k,t}(y)\|_2
&=\left\|x-y+(t-t_k)\big(\widehat u(x,t_k,s,a)-\widehat u(y,t_k,s,a)\big)\right\|_2 \\
&\ge \|x-y\|_2-(t-t_k)\|\widehat u(x,t_k,s,a)-\widehat u(y,t_k,s,a)\|_2 \\
&\ge \left(1-\frac{C(t-t_k)(1+L_\tau)}{\tau}\right)\|x-y\|_2,
\end{aligned}
\]
Since $L_\tau=\cO(\tau^{-1})$ and $\Delta=o(\tau^2)$, for sufficiently small $\tau$, we have $\frac{C\Delta(1+L_\tau)}{\tau}\le \frac{2C\Delta}{\tau^2}\le \frac12.$
Consequently,
\[
\|\Phi_{k,t}(x)-\Phi_{k,t}(y)\|_2\ge \frac12\|x-y\|_2,~~ x,y\in\mathbb R^{d_s}.
\]
Thus $\Phi_{k,t}$ is injective. We next prove that $\Phi_{k,t}$ is surjective. Fix any $z\in\mathbb R^{d_s}$. It suffices to find $x\in\mathbb R^{d_s}$ such that $\Phi_{k,t}(x)=z$. Since
\[
\Phi_{k,t}(x)=x+(t-t_k)\widehat u(x,t_k,s,a),
\]
the equation $\Phi_{k,t}(x)=z$ is equivalent to
\[
x=z-(t-t_k)\widehat u(x,t_k,s,a).
\]
For this fixed $z$, define
\[
T_z(x)=z-(t-t_k)\widehat u(x,t_k,s,a).
\]
Then, for any $x,y\in\mathbb R^{d_s}$,
\[
\begin{aligned}
\|T_z(x)-T_z(y)\|_2
&=(t-t_k)\|\widehat u(x,t_k,s,a)-\widehat u(y,t_k,s,a)\|_2 \\
&\le \frac{C(t-t_k)(1+L_\tau)}{\tau}\|x-y\|_2 \\
&\le \frac{1}{2}\|x-y\|_2 .
\end{aligned}
\]
Thus $T_z$ is a contraction on $\mathbb R^{d_s}$. Then, there exists a unique $x\in\mathbb R^{d_s}$ such that $T_z(x)=x$. This fixed point equation is exactly equivalent to $\Phi_{k,t}(x)=z$. Therefore $\Phi_{k,t}$ is surjective and hence bijective. Since
\[
\nabla_x\Phi_{k,t}(x)=I_{d_s}+(t-t_k)\nabla_x\widehat u(x,t_k,s,a),
\]
all singular values of $\nabla_x\Phi_{k,t}(x)$ are bounded below by $1-\frac{C(t-t_k)(1+L_\tau)}{\tau}\ge \frac{1}{2}.$ Therefore $\nabla_x\Phi_{k,t}(x)$ is invertible everywhere. Since $\widehat u$ is $C^2$ in $x$, the inverse function theorem implies that $\Phi_{k,t}$ is a diffeomorphism.	For $t\in[t_k,t_{k+1}]$, the EM interpolation satisfies
\[
\widehat X_t=\widehat X_{t_k}+(t-t_k)\widehat u(\widehat X_{t_k},t_k,s,a)=\Phi_{k,t}(\widehat X_{t_k}).
\]
Thus by the Liouville equation for deterministic flows, the density $\widehat p_t(\cdot|  s,a)$ satisfies
\[
\partial_t\widehat p_t(x|  s,a)+\nabla_x \cdot \Big[\widehat p_t(x|  s,a)\widehat u(\Phi_{k,t}^{-1}(x),t_k,s,a)\Big]=0.
\]
The true density satisfies
\[
\partial_t p_t(x|  s,a)+\nabla_x \cdot \Big[p_t(x|  s,a)u^*(x,t,s,a)\Big]=0.
\]
The Gaussian initialization in Algorithm~\ref{algo:integrated-estimation} gives $\widehat p_\tau(\cdot|  s,a)=q_\tau$, so the initial density difference is $q_\tau-p_\tau(\cdot|  s,a)$. Subtracting the two transport equations gives
\begin{align*}
&\partial_t\big(\widehat p_t(x|  s,a)-p_t(x|  s,a)\big)+\nabla_x \cdot \Big[\big(\widehat p_t(x|  s,a)-p_t(x|  s,a)\big)\widehat u(\Phi_{k,t}^{-1}(x),t_k,s,a)\Big]\\
&=-\nabla_x \cdot \Big[p_t(x|  s,a)\big(\widehat u(\Phi_{k,t}^{-1}(x),t_k,s,a)-u^*(x,t,s,a)\big)\Big].
\end{align*}
The $L^1$ stability estimate gives
\begin{align*}
&\Big\|\widehat p_{t_K}(\cdot |  s,a)-p_{1-\tau}(\cdot |  s,a) \Big\|_1 \\
&\le \|q_\tau-p_\tau(\cdot|  s,a)\|_1 +\int_\tau^{1-\tau}\left\|\nabla_x \cdot \Big[p_t(\cdot |  s,a)\Big(\widehat u(\Phi_{k,t}^{-1}(\cdot),t_k,s,a)-u^*(\cdot,t,s,a)\Big)\Big]\right\|_1dt\\
& \leq \underbrace{\|q_\tau-p_\tau(\cdot|  s,a)\|_1}_{\mbox{Term}~\textbf{I}}+\underbrace{ \int_\tau^{1-\tau}\left\|\nabla_x \cdot \Big(p_t(\cdot |  s,a)\big(\widehat u(\cdot,t,s,a)-u^*(\cdot,t,s,a)\big)\Big)\right\|_1dt 
}_{\mbox{Term}~\textbf{II}} \\
&~~~~ + \underbrace{ \int_\tau^{1-\tau}\left\|\nabla_x \cdot \Big(p_t(\cdot|  s,a)\big(\widehat{u}(\Phi_{k,t}^{-1}(\cdot),t_k,s,a)-\widehat u(\cdot,t,s,a)\big)\Big)\right\|_1dt 
}_{\mbox{Term}~\textbf{III}}.
\end{align*}

\paragraph{Bound Term I.}
Write $\sigma_\tau^2=1-\tau$. The learned flow starts from $q_\tau$, whereas the exact interpolating density at time $\tau$ is $p_\tau(\cdot|  s,a)$. Conditional on $X$, the interpolating variable $X_\tau=\tau X+\sigma_\tau\xi$ has distribution $\mathcal N(\tau X,\sigma_\tau^2I_{d_s})$. Thus $p_\tau(\cdot|  s,a)$ is a mixture of these Gaussian densities. By convexity of relative entropy in its first argument and the formula for relative entropy between Gaussians with the same covariance,
\[
\begin{aligned}
\operatorname{KL}\big(p_\tau(\cdot|  s,a)\|q_\tau\big)
&\le \mathbb E\left[\operatorname{KL}\big(\mathcal N(\tau X,\sigma_\tau^2I_{d_s})\|\mathcal N(0,\sigma_\tau^2I_{d_s})\big)|  s,a\right]\\
&=\frac{\tau^2}{2(1-\tau)}\mathbb E\big[\|X\|_2^2|  s,a\big].
\end{aligned}
\]
Using Pinsker's inequality, we obtain
\[
\TV^2\big(p_\tau(\cdot|  s,a),q_\tau\big)\le \frac12\operatorname{KL}\big(p_\tau(\cdot|  s,a)\|q_\tau\big)\le \frac{\tau^2}{4(1-\tau)}\mathbb E\big[\|X\|_2^2|  s,a\big].
\]
Since $\|X\|_2^2\le d_s$ by Assumption~\ref{ass:conditional-support}, this implies the initial $L^1$ bound
\[
\|q_\tau-p_\tau(\cdot|  s,a)\|_1^2=4\TV^2\big(p_\tau(\cdot|  s,a),q_\tau\big)\lesssim \frac{\tau^2}{1-\tau}.
\]
\paragraph{Bound Term II.}
Since
\[
\widehat u(x,t,s,a)-u^*(x,t,s,a)=\frac{2-t}{2t}\Big(s(x,t,s,a)-\nabla_x\log p_t(x|  s,a)\Big),
\]
we have
\begin{align*}
&\int_\tau^{1-\tau}\left\|\nabla_x \cdot \Big(p_t(\cdot|  s,a)\big(\widehat u(\cdot,t,s,a)-u^*(\cdot,t,s,a)\big)\Big)\right\|_1dt\\
&\le \frac{2-\tau}{2\tau}\int_\tau^{1-\tau}\left\|\nabla_x \cdot \Big(p_t(\cdot|  s,a)\big(s(\cdot,t,s,a)-\nabla_x\log p_t(\cdot|  s,a)\big)\Big)\right\|_1dt.
\end{align*}
By the time integrated Gagliardo--Nirenberg interpolation inequality,
\[
\begin{aligned}
&\int_\tau^{1-\tau}\left\|\nabla_x \cdot\Big[p_t(\cdot|  s,a)\big(s(\cdot,t,s,a)-\nabla_x\log p_t(\cdot|  s,a)\big)\Big]\right\|_1dt \\
&\lesssim \left(\int_\tau^{1-\tau}\int_{\mathbb R^{d_s}}p_t(x|  s,a)\left\|s(x,t,s,a)-\nabla_x\log p_t(x|  s,a)\right\|_2dx dt\right)^{\frac12} \\
&~~\times \left(\int_\tau^{1-\tau}\int_{\mathbb R^{d_s}}\left\|\nabla_x\Big[\nabla_x \cdot \Big(p_t(x|  s,a)\big(s(x,t,s,a)-\nabla_x\log p_t(x|  s,a)\big)\Big)\Big]\right\|_2dxdt\right)^{\frac12}.
\end{aligned}
\]
For the first factor,
\[
\begin{aligned}
&\int_{\mathbb R^{d_s}}p_t(x|  s,a)\left\|s(x,t,s,a)-\nabla_x\log p_t(x|  s,a)\right\|_2dx \\
&\le \left(\int_{\mathbb R^{d_s}}p_t(x|  s,a)dx\right)^{\frac12}\left(\int_{\mathbb R^{d_s}}p_t(x|  s,a)\left\|s(x,t,s,a)-\nabla_x\log p_t(x|  s,a)\right\|_2^2dx\right)^{\frac12} \\
&=\left(\mathbb E_{x_t\sim p_t(\cdot|  s,a)}\left\|s(x_t,t,s,a)-\nabla_{x_t}\log p_t(x_t|  s,a)\right\|_2^2\right)^{\frac12}.
\end{aligned}
\]
Thus,
\begin{align*}
&\int_\tau^{1-\tau}\int_{\mathbb R^{d_s}}p_t(x|  s,a)\left\|s(x,t,s,a)-\nabla_x\log p_t(x|  s,a)\right\|_2dxdt \\
&\le (1-2\tau)^{\frac12}\left(\int_\tau^{1-\tau}\mathbb E_{x_t\sim p_t(\cdot|  s,a)}\left\|s(x_t,t,s,a)-\nabla_{x_t}\log p_t(x_t|  s,a)\right\|_2^2dt\right)^{\frac12}.
\end{align*}
For the second factor, using
\[
p_t(x|  s,a)\big(s(x,t,s,a)-\nabla_x\log p_t(x|  s,a)\big)=p_t(x|  s,a)s(x,t,s,a)-\nabla_xp_t(x|  s,a),
\]
we have
\[
\begin{aligned}
	&\int_\tau^{1-\tau}\int_{\mathbb R^{d_s}}\Big\|\nabla_x\Big[\nabla_x\cdot\Big(p_t(x|  s,a)\big(s(x,t,s,a)-\nabla_x\log p_t(x|  s,a)\big)\Big)\Big]\Big\|_2dxdt \\
	&\le \int_\tau^{1-\tau}\int_{\mathbb R^{d_s}}\Big\|\nabla_x\Big(\nabla_x\cdot\big(p_t(x|  s,a)s(x,t,s,a)\big)\Big)\Big\|_2dxdt\\
    &~~~~ +\int_\tau^{1-\tau}\int_{\mathbb R^{d_s}}\Big\|\nabla_x\big(\nabla_x\cdot\nabla_xp_t(x|  s,a)\big)\Big\|_2dxdt .
\end{aligned}
\]
Moreover, for each $i=1,\ldots,d_s$, we have
\begin{align*}
\left[\nabla_x\Big(\nabla_x \cdot\big(p_t(x|  s,a)s(x,t,s,a)\big)\Big)\right]^{(i)}
&=\partial_{x_i}\Big[\sum_{j=1}^{d_s}\partial_{x_j}\big(p_t(x|  s,a)s^{(j)}(x,t,s,a)\big)\Big]\\
&=\sum_{j=1}^{d_s}\partial_{x_i}\partial_{x_j}\Big(p_t(x|  s,a)s^{(j)}(x,t,s,a)\Big).
\end{align*}
By the product rule,
\[
\begin{aligned}
\partial_{x_i}\partial_{x_j}\Big(p_t(x|  s,a)s^{(j)}(x,t,s,a)\Big)
&=\big(\partial_{x_i}\partial_{x_j}p_t(x|  s,a)\big)s^{(j)}(x,t,s,a) \\
&~~+\big(\partial_{x_j}p_t(x|  s,a)\big)\big(\partial_{x_i}s^{(j)}(x,t,s,a)\big) \\
&~~+\big(\partial_{x_i}p_t(x|  s,a)\big)(\partial_{x_j}s^{(j)}(x,t,s,a)) \\
&~~ +p_t(x|  s,a)\big(\partial_{x_i}\partial_{x_j}s^{(j)}(x,t,s,a)\big).
\end{aligned}
\]
To bound the second factor, we first record several elementary bounds for $p_t$. By Assumption~\ref{ass:conditional-support}, for $t\in[\tau,1-\tau]$,
\[
p_t(x|  s,a)=\int_{[0,1]^{d_s}}\frac{p_1(y|  s,a)}{(2\pi(1-t))^{d_s/2}} \exp\left(-\frac{\|x-ty\|_2^2}{2(1-t)}\right)dy.
\]
Differentiating the Gaussian kernel gives
\[
\nabla_x p_t(x|  s,a)=-\int_{[0,1]^{d_s}}\frac{x-ty}{1-t}\frac{p_1(y|  s,a)}{(2\pi(1-t))^{d_s/2}}\exp\left(-\frac{\|x-ty\|_2^2}{2(1-t)}\right)dy,
\]
and
\[
\nabla_x^2p_t(x|  s,a)=\int_{[0,1]^{d_s}}\left(\frac{(x-ty)(x-ty)^\top}{(1-t)^2}-\frac{I_{d_s}}{1-t}\right)\frac{p_1(y|  s,a) }{(2\pi(1-t))^{d_s/2}}\exp\left(-\frac{\|x-ty\|_2^2}{2(1-t)}\right)dy.
\]
Moreover, differentiating the Gaussian kernel three times gives
\[
\left \|\nabla_x\big[\nabla_x\cdot\nabla_xp_t(x|  s,a)\big]\right\|_2\lesssim \int_{[0,1]^{d_s}}\frac{1+\left\|\frac{x-ty}{\sqrt{1-t}}\right\|_2^3}{(1-t)^{3/2}}\frac{p_1(y|  s,a)}{(2\pi(1-t))^{d_s/2}}\exp\left(-\frac{\|x-ty\|_2^2}{2(1-t)}\right)dy.
\]
Using the change of variables $x=ty+\sqrt{1-t}z$ and the fact that $y\in[0,1]^{d_s}$, we obtain
\[
\int_{\mathbb R^{d_s}}\|\nabla_xp_t(x|  s,a)\|_2dx\lesssim \frac{1}{\sqrt{1-t}},
\]
\[
\int_{\mathbb R^{d_s}}(1+\|x\|_2)\|\nabla_x^2p_t(x|  s,a)\|_2dx\lesssim\frac{1}{1-t},
\]
and
\[
\int_{\mathbb R^{d_s}}\left \|\nabla_x\big[\nabla_x\cdot\nabla_xp_t(x|  s,a)\big] \right\|_2dx\lesssim \frac{1}{(1-t)^{3/2}}.
\]
Also, since $p_t(\cdot|  s,a)$ is a probability density and $X_t=tX_1+\sqrt{1-t}\xi$ with $X_1\in[0,1]^{d_s}$,
\[
\int_{\mathbb R^{d_s}}p_t(x|  s,a)dx=1,~~ \int_{\mathbb R^{d_s}}(1+\|x\|_2)p_t(x|  s,a)dx\lesssim 1.
\]
Therefore, using
\[
p_t(x|  s,a)\big(s(x,t,s,a)-\nabla_x\log p_t(x|  s,a)\big)=p_t(x|  s,a)s(x,t,s,a)-\nabla_xp_t(x|  s,a),
\]
we have
\[
\begin{aligned}
&\int_\tau^{1-\tau}\int_{\mathbb R^{d_s}}\left\|\nabla_x\Big[\nabla_x\cdot \Big(p_t(x|  s,a)\big(s(x,t,s,a)-\nabla_x\log p_t(x|  s,a)\big)\Big)\Big]\right\|_2dxdt \\
&\le \int_\tau^{1-\tau}\int_{\mathbb R^{d_s}}\left\|\nabla_x\Big(\nabla_x\cdot \big(p_t(x|  s,a)s(x,t,s,a)\big)\Big)\right\|_2dxdt\\
&~~~~ +\int_\tau^{1-\tau}\int_{\mathbb R^{d_s}}\left \|\nabla_x\big[\nabla_x\cdot\nabla_xp_t(x|  s,a)\big] \right\|_2dxdt \\
&\lesssim L_\tau\int_\tau^{1-\tau}\int_{\mathbb R^{d_s}}\Big((1+\|x\|_2)\|\nabla_x^2p_t(x|  s,a)\|_2+\|\nabla_xp_t(x|  s,a)\|_2+p_t(x|  s,a)\Big)dxdt \\
&~~+\int_\tau^{1-\tau}\int_{\mathbb R^{d_s}}\left \|\nabla_x\big[\nabla_x\cdot\nabla_xp_t(x|  s,a)\big] \right\|_2dxdt \\
&\lesssim \tau^{-1}\int_\tau^{1-\tau}\left(\frac{1}{1-t}+\frac{1}{\sqrt{1-t}}+1\right)dt+\int_\tau^{1-\tau}\frac{1}{(1-t)^{3/2}}dt \\
&\lesssim \frac{1}{\tau^2}.
\end{aligned}
\]
Combining the two factors gives
\begin{equation}\label{eq:tv-1}
\begin{aligned}
&\int_\tau^{1-\tau}\left\|\nabla_x \cdot \Big(p_t(\cdot |  s,a)\big(\widehat u(\cdot,t,s,a)-u^*(\cdot,t,s,a)\big)\Big)\right\|_1dt \\
&\lesssim \frac{1}{\tau^2} \left(\int_\tau^{1-\tau}\mathbb E_{x_t\sim p_t(\cdot |  s,a)}\left\|s(x_t,t,s,a)-\nabla_{x_t}\log p_t(x_t|  s,a)\right\|_2^2dt\right)^{\frac14}.
\end{aligned}
\end{equation}

\paragraph{Bound Term III.}
We compare the frozen EM velocity with the continuous velocity. Here $\nabla_1\widehat u$ denotes the derivative of $\widehat u$ with respect to its first argument. Since
\[
x=\Phi_{k,t}(\Phi_{k,t}^{-1}(x))=\Phi_{k,t}^{-1}(x)+(t-t_k)\widehat u(\Phi_{k,t}^{-1}(x),t_k,s,a),
\]
we first control $\widehat u(\Phi_{k,t}^{-1}(x),t_k,s,a)$. By the definition of $\widehat u$ and Assumption~\ref{ass:score-regularity},
\begin{equation}\label{eq:tv-eq-1}
\begin{aligned}
\|\widehat u(\Phi_{k,t}^{-1}(x),t_k,s,a)\|_2
&=\left\|\frac{\Phi_{k,t}^{-1}(x)}{t_k}+\frac{2-t_k}{2t_k}s(\Phi_{k,t}^{-1}(x),t_k,s,a)\right\|_2 \\
&\le \frac{\|\Phi_{k,t}^{-1}(x)\|_2}{t_k}+\frac{2-t_k}{2t_k}\|s(\Phi_{k,t}^{-1}(x),t_k,s,a)\|_2 \\
&\le \frac{1}{\tau}\|\Phi_{k,t}^{-1}(x)\|_2+\frac{L_\tau}{\tau}\Big(1+\|\Phi_{k,t}^{-1}(x)\|_2\Big) \\
&\lesssim \frac{1}{\tau^2}\Big(1+\|\Phi_{k,t}^{-1}(x)\|_2\Big).
\end{aligned}
\end{equation}
Since
\[
\Phi_{k,t}^{-1}(x)=x-(t-t_k)\widehat u(\Phi_{k,t}^{-1}(x),t_k,s,a),
\]
we have
\begin{align*}
1+\|\Phi_{k,t}^{-1}(x)\|_2&\le 1+\|x\|_2+(t-t_k)\|\widehat u(\Phi_{k,t}^{-1}(x),t_k,s,a)\|_2\\
&\lesssim 1+\|x\|_2+\frac{\Delta}{\tau^2}\Big(1+\|\Phi_{k,t}^{-1}(x)\|_2\Big).
\end{align*}
Then, we obtain
\[
1+\|\Phi_{k,t}^{-1}(x)\|_2\le 2\Big(1+\|x\|_2\Big).
\]
Substituting this bound into \eqref{eq:tv-eq-1} gives
\[
\|\widehat u(\Phi_{k,t}^{-1}(x),t_k,s,a)\|_2\lesssim \frac{1}{\tau^2}\Big(1+\|x\|_2\Big).
\]
Thus,
\[
\|x-\Phi_{k,t}^{-1}(x)\|_2=(t-t_k)\|\widehat u(\Phi_{k,t}^{-1}(x),t_k,s,a)\|_2\lesssim \frac{\Delta}{\tau^2}\Big(1+\|x\|_2\Big).
\]
Next, by the mean value theorem in $x$ and $t$,
\[
\begin{aligned}
&\|\widehat u(\Phi_{k,t}^{-1}(x),t_k,s,a)-\widehat u(x,t,s,a)\|_2 \\
&\le \|\widehat u(\Phi_{k,t}^{-1}(x),t_k,s,a)-\widehat u(x,t_k,s,a)\|_2+\|\widehat u(x,t_k,s,a)-\widehat u(x,t,s,a)\|_2 \\
&\lesssim \frac{1}{\tau^2}\|x-\Phi_{k,t}^{-1}(x)\|_2+\int_{t_k}^{t}\|\partial_r\widehat u(x,r,s,a)\|_2dr \\
&\lesssim \frac{\Delta}{\tau^4}\Big(1+\|x\|_2\Big).
\end{aligned}
\]
Here we used
\[
\|\nabla_1\widehat u(z,r,s,a)\|_2\lesssim \frac{1}{\tau^2},~~ \|\partial_r\widehat u(x,r,s,a)\|_2\lesssim \frac{1}{\tau^3}\Big(1+\|x\|_2\Big),
\]
which follow from the definition of $\widehat u$ and Assumption~\ref{ass:score-regularity}. For the derivative term, by the chain rule,
\[
\nabla_x\Big[\widehat u(\Phi_{k,t}^{-1}(x),t_k,s,a)\Big]=\nabla_1\widehat u(\Phi_{k,t}^{-1}(x),t_k,s,a)\nabla_x\Phi_{k,t}^{-1}(x).
\]
By \eqref{eq:tv-fun}, we have
\[
\nabla_z\Phi_{k,t}(z)=I_{d_s}+(t-t_k)\nabla_1\widehat u(z,t_k,s,a).
\]
Taking $z=\Phi_{k,t}^{-1}(x)$ and differentiating the identity
\[
\Phi_{k,t}(\Phi_{k,t}^{-1}(x))=x,
\]
we obtain
\[
\Big(I_{d_s}+(t-t_k)\nabla_1\widehat u(\Phi_{k,t}^{-1}(x),t_k,s,a)\Big)\nabla_x\Phi_{k,t}^{-1}(x)=I_{d_s}.
\]
Therefore,
\[
\nabla_x\Phi_{k,t}^{-1}(x)=\Big(I_{d_s}+(t-t_k)\nabla_1\widehat u(\Phi_{k,t}^{-1}(x),t_k,s,a)\Big)^{-1}.
\]
Since
\[
\left\|(t-t_k)\nabla_1\widehat u(\Phi_{k,t}^{-1}(x),t_k,s,a)\right\|_2\lesssim (t-t_k)\frac{(1+L_\tau)}{\tau}\lesssim \frac{\Delta}{\tau^2}\le \frac12,
\]
we have
\[
\begin{aligned}
\|\nabla_x\Phi_{k,t}^{-1}(x)\|_2
&=\left\|\Big(I_{d_s}+(t-t_k)\nabla_1\widehat u(\Phi_{k,t}^{-1}(x),t_k,s,a)\Big)^{-1}\right\|_2 \\
&\le \frac{1}{1-\left\|(t-t_k)\nabla_1\widehat u(\Phi_{k,t}^{-1}(x),t_k,s,a)\right\|_2}\le 2.
\end{aligned}
\]
Moreover,
\[
\begin{aligned}
\nabla_x\Phi_{k,t}^{-1}(x)-I_{d_s}
&=\Big(I_{d_s}+(t-t_k)\nabla_1\widehat u(\Phi_{k,t}^{-1}(x),t_k,s,a)\Big)^{-1}-I_{d_s} \\
&=-\Big(I_{d_s}+(t-t_k)\nabla_1\widehat u(\Phi_{k,t}^{-1}(x),t_k,s,a)\Big)^{-1}(t-t_k)\nabla_1\widehat u(\Phi_{k,t}^{-1}(x),t_k,s,a).
\end{aligned}
\]
Therefore,
\[
\|\nabla_x\Phi_{k,t}^{-1}(x)-I_{d_s}\|_2\le \frac{(t-t_k)\left\|\nabla_1\widehat u(\Phi_{k,t}^{-1}(x),t_k,s,a)\right\|_2}{1-\left\|(t-t_k)\nabla_1\widehat u(\Phi_{k,t}^{-1}(x),t_k,s,a)\right\|_2} \lesssim \frac{\Delta}{\tau^2}.
\]
Therefore, by the mean value theorem and Assumption~\ref{ass:score-regularity},
\[
\begin{aligned}
&\left\|\nabla_x\Big[\widehat u(\Phi_{k,t}^{-1}(x),t_k,s,a)\Big]-\nabla_1\widehat u(x,t,s,a)\right\|_2 \\
&=\left\|\nabla_1\widehat u(\Phi_{k,t}^{-1}(x),t_k,s,a)\nabla_x\Phi_{k,t}^{-1}(x)-\nabla_1\widehat u(x,t,s,a)\right\|_2 \\
&\le \|\nabla_1\widehat u(\Phi_{k,t}^{-1}(x),t_k,s,a)-\nabla_1\widehat u(x,t,s,a)\|_2\|\nabla_x\Phi_{k,t}^{-1}(x)\|_2 \\
&~~+\|\nabla_1\widehat u(x,t,s,a)\|_2\|\nabla_x\Phi_{k,t}^{-1}(x)-I_{d_s}\|_2 \\
&\lesssim \frac{\Delta}{\tau^4}\Big(1+\|x\|_2\Big).
\end{aligned}
\]
By the product rule,
\[
\begin{aligned}
&\left\|\nabla_x\cdot \Big(p_t(\cdot |  s,a)\big(\widehat u(\Phi_{k,t}^{-1}(\cdot ),t_k,s,a)-\widehat u(\cdot ,t,s,a)\big)\Big)\right\|_1 \\
&\le \int_{\mathbb R^{d_s}}\|\nabla_xp_t(x|  s,a)\|_2 \cdot \|\widehat u(\Phi_{k,t}^{-1}(x),t_k,s,a)-\widehat u(x,t,s,a)\|_2dx \\
&~~+\int_{\mathbb R^{d_s}}p_t(x|  s,a) \cdot \left\|\nabla_x\big(\widehat u(\Phi_{k,t}^{-1}(x),t_k,s,a)\big)-\nabla_1\widehat u(x,t,s,a)\right\|_2dx \\
&\lesssim \frac{\Delta}{\tau^4}\int_{\mathbb R^{d_s}}\Big(1+\|x\|_2\Big)\Big(\|\nabla_xp_t(x|  s,a)\|_2+p_t(x|  s,a)\Big)dx.
\end{aligned}
\]
Thus,
\begin{equation}\label{eq:tv-2}
\begin{aligned}
\int_\tau^{1-\tau}\left\|\nabla_x\cdot \Big[p_t(\cdot|  s,a)\Big(\widehat u(\Phi_{k,t}^{-1}(\cdot),t_k,s,a)-\widehat u(\cdot,t,s,a)\Big)\Big]\right\|_1dt &\lesssim \frac{\Delta}{\tau^4}\int_\tau^{1-\tau}\frac{1}{1-t}dt\\ 
&\lesssim \frac{\Delta}{\tau^4}\log\frac1\tau.
\end{aligned}
\end{equation}
Combining \eqref{eq:tv-1} and \eqref{eq:tv-2} with the preceding estimate, we obtain
\begin{align*}
\|\widehat p_{t_K}(\cdot|  s,a)-p_{1-\tau}(\cdot|  s,a)\|_1
&\lesssim \|q_\tau-p_\tau(\cdot|  s,a)\|_1 + \frac{\Delta}{\tau^4}\log\frac1\tau \\
&+\frac{1}{\tau^2} \left(\int_\tau^{1-\tau}\mathbb E_{x_t\sim p_t(\cdot|  s,a)}\left\|s(x_t,t,s,a)-\nabla_{x_t}\log p_t(x_t|  s,a)\right\|_2^2dt\right)^{\frac14}.
\end{align*}
Taking expectation over $\bD_x,\bT,\bZ$ and $(S,A)$, and using Assumption~\ref{ass:score-estimate} together with the initial $L^1$ bound established above, gives
\[
\begin{aligned}
\mathbb E_{\bD_x,\bT,\bZ}\mathbb E_{(S,A)}\|\widehat p_{t_K}(\cdot|  S,A)-p_{1-\tau}(\cdot|  S,A)\|_1 \lesssim \tau\sqrt{\frac{1}{1-\tau}}+\frac{\zeta^{\frac12}}{\tau^2}  + \frac{\Delta}{\tau^4}\log\frac1\tau.
\end{aligned}
\]
Finally, by the definition of total variation distance, for every measurable set $D\subseteq\mathbb R^{d_s}$,
\[
\left|\int_D\widehat p_{t_K}(x|  S,A)dx-\int_Dp_{1-\tau}(x|  S,A)dx\right|\le \int_{\mathbb R^{d_s}}\left|\widehat p_{t_K}(x|  S,A)-p_{1-\tau}(x|  S,A)\right|dx.
\]
Taking the supremum over all measurable $D$ yields
\[
\TV^2\Big(\widehat p_{t_K}(\cdot|  S,A),p_{1-\tau}(\cdot|  S,A)\Big)\le \|\widehat p_{t_K}(\cdot|  S,A)-p_{1-\tau}(\cdot|  S,A)\|^2_1.
\]
Therefore,
\begin{align*}
\TV^2\Big(\widehat p_{t_K}&(\cdot|  S,A),p_{1-\tau}(\cdot|  S,A)\Big)\lesssim\|q_\tau-p_\tau(\cdot|  S,A)\|_1^2 +\frac{\Delta^2}{\tau^8}\log^2 \tau\\
&+\frac{1}{\tau^4}\left(\int_\tau^{1-\tau}\mathbb E_{x_t\sim p_t(\cdot|  S,A)}\left\|s(x_t,t,S,A)-\nabla_{x_t}\log p_t(x_t|  S,A)\right\|_2^2dt\right)^{\frac12}.
\end{align*}
Taking expectations and using the initial $L^1$ bound established above, we have
\[
\mathbb E_{\bD_x,\bT,\bZ}\mathbb E_{(S,A)}\TV^2\Big(\widehat p_{t_K}(\cdot|  S,A),p_{1-\tau}(\cdot|  S,A)\Big) \lesssim\frac{\tau^2}{1-\tau}+\frac{\zeta}{\tau^4} + \frac{\Delta^2}{\tau^8}\log^2 \tau.
\]
This proves the claim.
\end{proof}
The preceding lemma controls the discrepancy due to Gaussian initialization, score estimation, and EM discretization up to time $1-\tau$. We next bound the remaining discrepancy between the exact flow at time $1-\tau$ and the terminal target distribution.

\subsection{Proof of Lemma \ref{lem:tv-time}}

\begin{proof}
Let $D=[0,1]^{d_s}$. Since $X_{1-\tau}=(1-\tau)X+\sqrt{\tau}\xi$, where $X\sim p_1(\cdot|  S,A)$, $\xi\sim\mathcal N(0,I_{d_s})$, and $p_1(\cdot|  S,A)$ is supported on $D$ by Assumption~\ref{ass:conditional-support}, the density of $X_{1-\tau}$ is
\[
p_{1-\tau}(y|  S,A)=\int_D\frac{1}{(2\pi\tau)^{\frac{d_s}{2}}}\exp\Big(-\frac{\|y-(1-\tau)x\|_2^2}{2\tau}\Big)p_1(x|  S,A)dx.
\]
Because $p_1(\cdot|  S,A)$ is supported on $D$,
\[
\int_{\mathbb R^{d_s}}\big|p_{1-\tau}(y|  S,A)-p_1(y|  S,A)\big|dy=\int_D\big|p_{1-\tau}(y|  S,A)-p_1(y|  S,A)\big|dy+\int_{D^c}p_{1-\tau}(y|  S,A)dy.
\]
We first bound the integral over $D$. For $y\in D$,
\[
\int_{\mathbb R^{d_s}}\frac{(1-\tau)^{d_s}}{(2\pi\tau)^{\frac{d_s}{2}}}\exp\Big(-\frac{\|y-(1-\tau)x\|_2^2}{2\tau}\Big)dx=1.
\]
Hence, by adding and subtracting $(1-\tau)^{d_s}p_{1-\tau}(y|  S,A)$,
\[
\begin{aligned}
&\int_D\big|p_{1-\tau}(y|  S,A)-p_1(y|  S,A)\big|dy \\
&\le \int_D\int_D\frac{(1-\tau)^{d_s}\big|p_1(y|  S,A)-p_1(x|  S,A)\big|}{(2\pi\tau)^{\frac{d_s}{2}}}\times \exp\Big(-\frac{\|y-(1-\tau)x\|_2^2}{2\tau}\Big)dxdy \\
&~~ + \big|1-(1-\tau)^{d_s}\big|\int_Dp_{1-\tau}(y|  S,A)dy  \\
&~~ + \int_Dp_1(y|  S,A)\int_{D^c}\frac{(1-\tau)^{d_s}}{(2\pi\tau)^{\frac{d_s}{2}}}\exp\Big(-\frac{\|y-(1-\tau)x\|_2^2}{2\tau}\Big)dxdy.
\end{aligned}
\]
For the first term, Assumption~\ref{ass:terminal-holder} implies
\[
\big|p_1(y|  S,A)-p_1(x|  S,A)\big|\lesssim \|y-x\|_2^{\beta\wedge1},~~ x,y\in D.
\]
For the first term, using $u=\frac{(1-\tau)x-y}{\sqrt{\tau}}$, namely $x=\frac{y+\sqrt{\tau}u}{1-\tau}$, we obtain
\[
\begin{aligned}
&\int_D\int_D\frac{(1-\tau)^{d_s}\big|p_1(y|  S,A)-p_1(x|  S,A)\big|}{(2\pi\tau)^{\frac{d_s}{2}}}\exp\Big(-\frac{\|y-(1-\tau)x\|_2^2}{2\tau}\Big)dxdy \\
&\lesssim  \int_D\int_D\frac{(1-\tau)^{d_s}\|y-x\|_2^{\beta\wedge1}}{(2\pi\tau)^{\frac{d_s}{2}}}\exp\Big(-\frac{\|y-(1-\tau)x\|_2^2}{2\tau}\Big)dxdy \\
&= \int_D\int_{\frac{(1-\tau)D-y}{\sqrt{\tau}}}\Big\|y-\frac{y+\sqrt{\tau}u}{1-\tau}\Big\|_2^{\beta\wedge1}(2\pi)^{-\frac{d_s}{2}}e^{-\frac{\|u\|_2^2}{2}}dudy \\
&\lesssim \int_D\int_{\mathbb R^{d_s}}\Big[\Big(\frac{\sqrt{\tau}}{1-\tau}\Big)^{\beta\wedge1}\|u\|_2^{\beta\wedge1}+\Big(\frac{\tau}{1-\tau}\Big)^{\beta\wedge1}\|y\|_2^{\beta\wedge1}\Big](2\pi)^{-\frac{d_s}{2}}e^{-\frac{\|u\|_2^2}{2}}dudy \\
&\lesssim \Big(\frac{\sqrt{\tau}}{1-\tau}\Big)^{\beta\wedge1}.
\end{aligned}
\]
The second term satisfies
\[
\big|1-(1-\tau)^{d_s}\big|\int_Dp_{1-\tau}(y|  S,A)dy\le \big|1-(1-\tau)^{d_s}\big|\lesssim \tau.
\]
For the third term, since $\|p_1(\cdot|  S,A)\|_\infty\le\Sigma$ by Assumption~\ref{ass:terminal-holder},
\[
\begin{aligned}
&\int_D p_1(y|  S,A)\int_{D^c}\frac{(1-\tau)^{d_s}}{(2\pi\tau)^{\frac{d_s}{2}}}\exp\Big(-\frac{\|y-(1-\tau)x\|_2^2}{2\tau}\Big)dxdy \\
&\lesssim \int_D\mathbb P\Big(\frac{y+\sqrt{\tau}\xi}{1-\tau}\notin D\Big)dy \\
&\lesssim \sum_{i=1}^{d_s}\int_D\Big[\mathbb P\Big(\xi_i<-\frac{y_i}{\sqrt{\tau}}\Big)+\mathbb P\Big(\xi_i>\frac{1-\tau-y_i}{\sqrt{\tau}}\Big)\Big]dy \\
&\lesssim \sum_{i=1}^{d_s}\int_0^1\Big[\Phi\Big(-\frac{s}{\sqrt{\tau}}\Big)+\Phi\Big(\frac{s-(1-\tau)}{\sqrt{\tau}}\Big)\Big]ds \\
&\lesssim  d_s\Big[\int_0^\infty\Phi\Big(-\frac{s}{\sqrt{\tau}}\Big)ds+\int_0^{1-\tau}\Phi\Big(-\frac{1-\tau-s}{\sqrt{\tau}}\Big)ds+\int_{1-\tau}^1 1ds\Big] \\
&\lesssim \sqrt{\tau}.
\end{aligned}
\]
Then, for sufficiently small $\tau$, we have
$$
\int_D\big|p_{1-\tau}(y|  S,A)-p_1(y|  S,A)\big|dy\lesssim \Big(\frac{\sqrt{\tau}}{1-\tau}\Big)^{\beta\wedge1}+ \tau + \Sigma \sqrt{\tau}\lesssim \Big(\frac{\sqrt{\tau}}{1-\tau}\Big)^{\beta\wedge1}.
$$
It remains to bound the integral over $D^c$. By the density formula and $\|p_1(\cdot|  S,A)\|_\infty\le\Sigma$,
\begin{align*}
\int_{D^c}p_{1-\tau}(y|  S,A)dy
&=\int_Dp_1(x|  S,A)\int_{D^c}\frac{1}{(2\pi\tau)^{\frac{d_s}{2}}}\exp\Big(-\frac{\|y-(1-\tau)x\|_2^2}{2\tau}\Big)dydx \\
&\lesssim \int_D\mathbb P\big((1-\tau)x+\sqrt{\tau}\xi\notin D\big)dx\\
& \lesssim  \sum_{i=1}^{d_s}\int_0^1\Big[\Phi\Big(-\frac{(1-\tau)s}{\sqrt{\tau}}\Big)+\Phi\Big(\frac{(1-\tau)s-1}{\sqrt{\tau}}\Big)\Big]ds\\
& \lesssim \frac{\sqrt{\tau}}{1-\tau}.
\end{align*}
Therefore, combining the preceding estimates, for sufficiently small $\tau$, we obtain
\begin{align*}
\TV\big(p_{1-\tau}(\cdot|  S,A),p_1(\cdot|  S,A)\big)
&\leq \int_{\mathbb R^{d_s}}\big|p_{1-\tau}(y|  S,A)-p_1(y|  S,A)\big|dy\\
&\lesssim \Sigma\Big(\frac{\sqrt{\tau}}{1-\tau}\Big)^{\beta\wedge1}+\frac{\sqrt{\tau}}{1-\tau}\lesssim \tau^{\frac{\beta\wedge1}{2}}.
\end{align*}
Hence
\[
\bE_{(S,A)}\Big[\TV^2\big(p_{1-\tau}(\cdot|  S,A),p_1(\cdot|  S,A)\big)\Big]\lesssim \tau^{\beta\wedge1}.
\]
This completes the proof of the lemma.
\end{proof}

Lemmas~\ref{lem:tv-score} and~\ref{lem:tv-time} control the two components of the total variation decomposition. We now combine them to obtain the full transition kernel bound and then transfer the same argument to the conditional reward law.

\subsection{Proofs of Lemma \ref{lem:TV-flow-x} and Corollary~\ref{cor:TV-flow-r}}
\begin{proof}[Proof of Lemma \ref{lem:TV-flow-x}]
By the data-processing inequality for total variation, the TV error of the clipped transition law $\widehat{\mathcal P}(\cdot|  S,A)$ is bounded by that of the flow output before clipping. Combining \eqref{eq:TV-decomposition} with Lemmas~\ref{lem:tv-score}--\ref{lem:tv-time} yields the desired bound. Here, the initialization term is absorbed into the terminal smoothing term, since $\frac{\tau^2}{1-\tau}\lesssim \tau^2\lesssim \tau^{\beta\wedge1}$ for sufficiently small $\tau$.
\end{proof}

Similarly, we obtain the following error bound for reward law estimation.
\begin{proof}[Proof of Corollary~\ref{cor:TV-flow-r}]
The proof follows the same argument as that of Lemma \ref{lem:TV-flow-x}, with the  next state variable replaced by the reward variable.
\end{proof}

Having established total variation bounds for both learned conditional laws, we next propagate these distributional errors, together with Monte-Carlo and importance sampling errors, through the learned soft Bellman operator.

\subsection{Proof of Theorem \ref{thmm:rl-flow-bound}}\label{sec:B.5}
\begin{proof}
By the definitions of $\cT^*_\alpha$ and $\widehat{\cT}_\alpha$, for any $Q$,
\begin{equation}\label{eq:bellman-1}
\begin{aligned}
&\Big(\cT^*_\alpha Q(S,A)-\widehat{\cT}_\alpha Q(S,A)\Big)^2\\
&\lesssim \left|\Ebb_{R\sim\cR(\cdot|  S,A)}[R]-\frac{1}{m}\sum_{j=1}^m\widehat R_j(S,A)\right|^2+\gamma^2\left|\Ebb_{S'\sim\cP(\cdot|  S,A)}V_Q(S')-\frac{1}{m}\sum_{j=1}^m{\widehat V}_Q(\widehat S'_j)\right|^2\\
&\lesssim \left|\Ebb_{R\sim\cR(\cdot|  S,A)}[R]-\Ebb_{R\sim\widehat\cR(\cdot|  S,A)}[R]\right|^2+\left|\Ebb_{R\sim\widehat\cR(\cdot|  S,A)}[R]-\frac{1}{m}\sum_{j=1}^m\widehat R_j(S,A)\right|^2\\
&~~+\gamma^2\left|\Ebb_{S'\sim\cP(\cdot|  S,A)}V_Q(S')-\Ebb_{S'\sim\widehat\cP(\cdot|  S,A)}V_Q(S')\right|^2\\
&~~ +\gamma^2\left|\Ebb_{S'\sim\widehat\cP(\cdot|  S,A)}V_Q(S')-\Ebb_{S'\sim\widehat\cP(\cdot|  S,A)}\widehat V_Q(S')\right|^2\\
&~~+\gamma^2\left|\Ebb_{S'\sim\widehat\cP(\cdot|  S,A)}{\widehat V}_Q(S')-\frac{1}{m}\sum_{j=1}^m{\widehat V}_Q(\widehat S'_j)\right|^2 .
\end{aligned}
\end{equation}
Fix $(S,A)$. Conditional on the learned reward law $\widehat{\mathcal R}$, the generated rewards are i.i.d. and bounded in $[0,R_{\max}]$. Hence
\begin{equation}\label{eq:bellman-2}
\begin{aligned}
\mathbb E_{\widehat R_{1:m}}\left[\left|\mathbb E_{R\sim\widehat{\mathcal R}(\cdot| S,A)}[R]-\frac{1}{m}\sum_{j=1}^m\widehat R_j(S,A)\right|^2 \middle| \widehat{\mathcal R}\right]=\frac{1}{m}\operatorname{Var}_{R\sim\widehat{\mathcal R}(\cdot| S,A)}(R)\le\frac{R_{\max}^2}{4m}.
\end{aligned}
\end{equation}
Moreover,
\begin{equation}\label{eq:bellman-3}
\begin{aligned}
\left|\Ebb_{R\sim\cR(\cdot|  S,A)}[R]-\Ebb_{R\sim\widehat\cR(\cdot|  S,A)}[R]\right|^2
&=\left|\int_{[0,R_{\max}]}R\Big[\cR(dR|  S,A)-\widehat\cR(dR|  S,A)\Big]\right|^2\\
&\lesssim R_{\max}^2\TV^2\Big(\cR(\cdot|  S,A),\widehat\cR(\cdot|  S,A)\Big).
\end{aligned}
\end{equation}
For the transition distribution term, since \(|V_Q(s')|\le Q_{\max}\), the integral characterization of total variation gives  
\begin{equation}\label{eq:transition-tv-common}
\begin{aligned}
\left|\Ebb_{S'\sim\cP(\cdot|  S,A)}V_Q(S')-\Ebb_{S'\sim\widehat\cP(\cdot|  S,A)}V_Q(S')\right|^2
&=\Big|\int_{[0,1]^{d_s}}V_Q(s')\big[\cP(ds'|  S,A)-\widehat\cP(ds'|  S,A)\big]\Big|^2\\
&\lesssim Q_{\max}^2\TV^2\Big(\cP(\cdot|  S,A),\widehat\cP(\cdot|  S,A)\Big).
\end{aligned}
\end{equation}
Suppose first that $Q=\widehat Q$. By Jensen's inequality and $\widehat Q\in\mathcal Q$,
\begin{align*}
\Ebb_{\bD_a}\Big|\Ebb_{S'\sim\widehat\cP(\cdot|  S,A)}\big[V_{\widehat Q}(S')-\widehat V_{\widehat Q}(S')\big]\Big|^2&\le\Ebb_{\bD_a}\Ebb_{S'\sim\widehat\cP(\cdot|  S,A)}
\left|V_{\widehat Q}(S')-\widehat V_{\widehat Q}(S')\right|^2\\
&\le\Ebb_{\bD_a}\sup_{Q\in\mathcal Q}\Ebb_{S'\sim\widehat\cP(\cdot|  S,A)}
\left|V_Q(S')-\widehat V_Q(S')\right|^2.
\end{align*}
Let $\mathcal Q_M$ be a $\frac{1}{M}$-cover of $\mathcal Q$ under $\|\cdot\|_\infty$. For every $Q\in\mathcal Q$, choose $Q_M\in\mathcal Q_M$ such that $\|Q-Q_M\|_\infty\le\frac{1}{M}$. The log-sum-exp inequality gives
\[
|V_Q(s')-V_{Q_M}(s')|\le\frac{1}{M},~~
|\widehat V_Q(s')-\widehat V_{Q_M}(s')|\le\frac{1}{M}.
\]
Hence,
\[
\begin{aligned}
|V_Q(s')-\widehat V_Q(s')|^2
&\le\left(\left|V_{Q_M}(s')-\widehat V_{Q_M}(s')\right|+\frac{2}{M}\right)^2\\
&\le2\left|V_{Q_M}(s')-\widehat V_{Q_M}(s')\right|^2+\frac{8}{M^2}.
\end{aligned}
\]
Taking the supremum over $Q\in\mathcal Q$ yields
\begin{equation}\label{eq:v-vhat}
\begin{aligned}
\sup_{Q\in\mathcal Q}|V_Q(s')-\widehat V_Q(s')|^2
\le2\max_{Q_M\in\mathcal Q_M}|V_{Q_M}(s')-\widehat V_{Q_M}(s')|^2+\frac{8}{M^2}.
\end{aligned}
\end{equation}
For every fixed $Q_M\in\mathcal Q_M$ and $s'$, the importance-sampling average is unbiased:
\[
\Ebb_{\bD_a}\left[\frac{1}{M}\sum_{j=1}^M\frac{\exp\left(\frac{Q_M(s',\widetilde A_j)}{\alpha}\right)}{f(\widetilde A_j)}\right]
=\int_{\mathcal A}\exp\left(\frac{Q_M(s',a)}{\alpha}\right)da.
\]
Moreover,
\[
\frac{\exp\left(-\frac{Q_{\max}}{\alpha}\right)}{c_2}
\le\frac{\exp\left(\frac{Q_M(s',\widetilde A_j)}{\alpha}\right)}{f(\widetilde A_j)}
\le\frac{\exp\left(\frac{Q_{\max}}{\alpha}\right)}{c_1}.
\]
Since
\[
\int_{\mathcal A}\exp\left(\frac{Q_M(s',a)}{\alpha}\right)da\ge\exp\left(-\frac{Q_{\max}}{\alpha}\right),~
\frac{1}{M}\sum_{j=1}^M\frac{\exp\left(\frac{Q_M(s',\widetilde A_j)}{\alpha}\right)}{f(\widetilde A_j)}
\ge\frac{\exp\left(-\frac{Q_{\max}}{\alpha}\right)}{c_2},
\]
the logarithm is Lipschitz on the interval containing both arguments, with Lipschitz constant at most $e^{\frac{\qm}{\alpha}}\max\{1,c_2\}$. Hoeffding's inequality therefore gives, for every $u>0$,
\[
\mathbb P_{\bD_a}\left(\left|V_{Q_M}(s')-\widehat V_{Q_M}(s')\right|\ge u\right)\le2\exp\left[-\frac{2Mu^2}{\alpha^2\max\{1,c_2^2\}\left(\frac{\exp\left(\frac{2Q_{\max}}{\alpha}\right)}{c_1}-\frac{1}{c_2}\right)^2}\right].
\]
Applying the union bound over the minimal cover $\mathcal Q_M$ gives, for every $t>0$,
\[
\begin{aligned}
&\mathbb P_{\bD_a}\left(\max_{Q_M\in\mathcal Q_M}|V_{Q_M}(s')-\widehat V_{Q_M}(s')|^2\ge t\right)\\
&\le 2\mathcal N\left(\mathcal Q,\frac{1}{M},\|\cdot\|_\infty\right)\exp\left[-\frac{2Mt}{\alpha^2\max\{1,c_2^2\}\left(\frac{\exp\left(\frac{2Q_{\max}}{\alpha}\right)}{c_1}-\frac{1}{c_2}\right)^2}\right],
\end{aligned}
\]
then
\[
\begin{aligned}
&\Ebb_{\bD_a}\max_{Q_M\in\mathcal Q_M}|V_{Q_M}(s')-\widehat V_{Q_M}(s')|^2\\
&=\int_0^\infty\mathbb P_{\bD_a}\left(\max_{Q_M\in\mathcal Q_M}|V_{Q_M}(s')-\widehat V_{Q_M}(s')|^2\ge t\right)dt\\
&\le\int_0^\infty\min\left\{1,2\mathcal N\left(\mathcal Q,\frac{1}{M},\|\cdot\|_\infty\right)\exp\left[-\frac{2Mt}{\alpha^2\max\{1,c_2^2\}\left(\frac{\exp(2Q_{\max}/\alpha)}{c_1}-\frac{1}{c_2}\right)^2}\right]\right\}dt\\
&\lesssim \frac{\exp(\frac{4\qm}{\alpha})}{M}
\log\left(2\mathcal N\big(\mathcal Q,\frac{1}{M},\|\cdot\|_\infty\big)\right).
\end{aligned}
\]
Combining Jensen's inequality, Fubini's theorem, and \eqref{eq:v-vhat}, we have
\begin{equation}\label{eq:qhat-action-error}
\begin{aligned}
&\Ebb_{\bD_a}\left|\Ebb_{S'\sim\widehat\cP(\cdot|  S,A)}\Big[V_{\widehat Q}(S')-\widehat V_{\widehat Q}(S')\Big]\right|^2\\
&\le\Ebb_{\bD_a}\sup_{Q\in\mathcal Q}\Ebb_{S'\sim\widehat\cP(\cdot|  S,A)}|V_Q(S')-\widehat V_Q(S')|^2\\
&\le\Ebb_{\bD_a}\Ebb_{S'\sim\widehat\cP(\cdot|  S,A)}\sup_{Q\in\mathcal Q}|V_Q(S')-\widehat V_Q(S')|^2\\
&\le\frac{8}{M^2}+2\Ebb_{S'\sim\widehat\cP(\cdot|  S,A)}\Ebb_{\bD_a}\max_{Q_M\in\mathcal Q_M}|V_{Q_M}(S')-\widehat V_{Q_M}(S')|^2\\
&\lesssim \frac{\exp(\frac{4\qm}{\alpha})}{M}\log\Big(2\mathcal N\big(\mathcal Q,\frac{1}{M},\|\cdot\|_\infty\big)\Big).
\end{aligned}
\end{equation}
Conditional on $\widehat\cP$ and $\bD_a$, $-Q_{\max}-\alpha\log c_2 \le\widehat V_Q(s') \le Q_{\max}-\alpha\log c_1.$
Thus, the range of $\widehat V_Q$ is bounded by \(2Q_{\max}+\alpha\log(c_2/c_1).\)
Let $\mathcal Q_m$ be a $\frac{1}{m}$-cover of $\mathcal Q$. Repeating the preceding covering argument and applying Hoeffding's inequality to the i.i.d. sample $\widehat S_1',\ldots,\widehat S_m'$ gives
\begin{equation}\label{eq:qhat-state-error}
\begin{aligned}
\Ebb_{\widehat S'_{1:m}}\left|\Ebb_{S'\sim\widehat\cP(\cdot|  S,A)}
\widehat V_{\widehat Q}(S')-\frac{1}{m}\sum_{i=1}^m\widehat V_{\widehat Q}(\widehat S_i')\right|^2\lesssim
\frac{\qm^2}{m}
\log\Big(2\mathcal N\big(\mathcal Q,\frac{1}{m},\|\cdot\|_\infty\big)\Big).
\end{aligned}
\end{equation}
For the ReLU DNN class $\mathcal Q$, choose
\[
\mathcal C=\cO(n^{\frac{d}{d+2\beta}}\log n),~~ \mathcal D=\cO(\log n),~~ \mathcal W=\cO(n^{\frac{d}{d+2\beta}}),~~ \mathcal B=\cO(n^{\frac{d}{d+2\beta}}).
\]
By Lemma \ref{lem:cover-nn}, the covering satisfies
\[
\log\mathcal N\left(\mathcal Q,\delta,\|\cdot\|_\infty\right)\lesssim\mathcal C\mathcal D\log\left(\frac{\mathcal B\mathcal W\mathcal D}{\delta}\right).
\]
Consequently,
\[
\log\mathcal N\left(\mathcal Q,\frac{1}{M},\|\cdot\|_\infty\right)\lesssim n^{\frac{d}{d+2\beta}}\log^2n\log (Mn),
\]
and
\[
\log\mathcal N\left(\mathcal Q,\frac{1}{m},\|\cdot\|_\infty\right)\lesssim n^{\frac{d}{d+2\beta}}\log^2n\log (mn).
\]
Thus by \eqref{eq:qhat-action-error} and \eqref{eq:qhat-state-error}, we obtain
\begin{equation}\label{eq:VQhat-VhatQhatmean}
\begin{aligned}
\Ebb_{\bD_a}\Big|\Ebb_{S'\sim\widehat\cP(\cdot|  S,A)}\big[V_{\widehat Q}(S')-\widehat V_{\widehat Q}(S')\big]\Big|^2&\lesssim  \frac{\exp(\frac{4\qm}{\alpha})}{M}n^{\frac{d}{d+2\beta}}\log^2n\log(Mn),\\
\Ebb_{\bD_x,\bT,\bZ}\Big|\Ebb_{S'\sim\widehat\cP(\cdot|  S,A)}
\widehat V_{\widehat Q}(S')-\frac{1}{m}\sum_{i=1}^m\widehat V_{\widehat Q}(\widehat S_i')\Big|^2&\lesssim \frac{Q_{\max}^2}{m}n^{\frac{d}{d+2\beta}}\log^2n\log(mn).
\end{aligned}
\end{equation}
Suppose now that $Q=Q^*$. Using $|\log x-\log y|^2\le |x-y|^2/(xy)$ for $x,y>0$, we obtain
\[
\begin{aligned}
&\Ebb_{\bD_a}\left|V_{Q^*}(s)-\widehat V_{Q^*}(s)\right|^2\\
&\le\frac{\alpha^2c_2}{\int_{\mathcal A}\exp\left(\frac{Q^*(s,a)}{\alpha}\right)da}\Ebb_{\bD_a}\Big|\frac{1}{M}\sum_{j=1}^M\frac{\exp\left(\frac{Q^*(s,\widetilde A_j)}{\alpha}\right)}{f(\widetilde A_j)}-\int_{\mathcal A}\exp\big(\frac{Q^*(s,a)}{\alpha}\big)da\Big|^2\\
&=\frac{\alpha^2c_2}{M\int_{\mathcal A}\exp\left(\frac{Q^*(s,a)}{\alpha}\right)da}\operatorname{Var}_{\widetilde A\sim f}\left(\frac{\exp\left(\frac{Q^*(s,\widetilde A)}{\alpha}\right)}{f(\widetilde A)}\right)\\
&\le\frac{\alpha^2c_2}{M\int_{\mathcal A}\exp\left(\frac{Q^*(s,a)}{\alpha}\right)da}\int_{\mathcal A}\frac{\exp\left(\frac{2Q^*(s,a)}{\alpha}\right)}{f(a)}da\\
&\lesssim \frac{\exp(\frac{\qm}{\alpha})}{M}.
\end{aligned}
\]
Consequently, Jensen's inequality and Fubini's theorem give
\begin{equation}\label{eq:qstar-action-error}
\begin{aligned}
\Ebb_{\bD_a}\left|\Ebb_{S'\sim\widehat\cP(\cdot|  S,A)}\Big[V_{Q^*}(S')-\widehat V_{Q^*}(S')\Big]\right|^2
&\le\Ebb_{\bD_a}\Ebb_{S'\sim\widehat\cP(\cdot|  S,A)}\left|V_{Q^*}(S')-\widehat V_{Q^*}(S')\right|^2\\
&=\Ebb_{S'\sim\widehat\cP(\cdot|  S,A)}\Ebb_{\bD_a}\left|V_{Q^*}(S')-\widehat V_{Q^*}(S')\right|^2\\
&\lesssim\frac{\exp\left(\frac{Q_{\max}}{\alpha}\right)}{M}.
\end{aligned}
\end{equation}
Conditional on the learned transition kernel $\widehat{\mathcal P}$ and the action sample $\mathbb D_a$, the generated next states are i.i.d. The preceding range bound yields
\begin{equation}\label{eq:qstar-state-error}
\begin{aligned}
\mathbb E_{\widehat S'_{1:m}}\left[\left|\mathbb E_{S'\sim\widehat{\mathcal P}(\cdot| S,A)}\widehat V_{Q^*}(S')-\frac{1}{m}\sum_{i=1}^m\widehat V_{Q^*}(\widehat S'_i)\right|^2 \middle| \widehat{\mathcal P},\mathbb D_a\right]=\frac{1}{m}\operatorname{Var}_{S'\sim\widehat{\mathcal P}(\cdot| S,A)}\left(\widehat V_{Q^*}(S')\right)\le\frac{Q_{\max}^2}{m}.
\end{aligned}
\end{equation}
Taking expectations over $\bD,\bT,\bZ$ and combining \eqref{eq:bellman-1}--\eqref{eq:transition-tv-common}, \eqref{eq:VQhat-VhatQhatmean}, and $\gamma<1$, we obtain
\begin{align*}
\bE_{\bD,\bT,\bZ}\left\|\cT^*_\alpha\widehat Q-\widehat{\cT}_\alpha\widehat Q\right\|_{L^2(\mu)}^2
&\lesssim R_{\max}^2\bE_{\bD,\bT,\bZ}\bE_{(S,A)}\left[\TV^2\big(\cR(\cdot|  S,A),\widehat\cR(\cdot|  S,A)\big)\right]\\
&~~+Q_{\max}^2\bE_{\bD,\bT,\bZ}\bE_{(S,A)}\left[\TV^2\big(\cP(\cdot|  S,A),\widehat\cP(\cdot|  S,A)\big)\right]\\
&~~+\frac{\exp\left(\frac{4Q_{\max}}{\alpha}\right)}{M}n^{\frac{d}{d+2\beta}}\log^2n\log(Mn)\\
&~~+\frac{Q_{\max}^2}{m}n^{\frac{d}{d+2\beta}}\log^2n\log(mn).
\end{align*}
Combining \eqref{eq:bellman-1}--\eqref{eq:transition-tv-common}, \eqref{eq:qstar-action-error}--\eqref{eq:qstar-state-error}, and using $R_{\max}\le Q_{\max}$ and $\gamma<1$, we obtain
\begin{align*}
\bE_{\bD,\bT,\bZ}\left\|\widehat{\cT}_\alpha Q^*-\cT^*_\alpha Q^*\right\|_{L^2(\mu)}^2
&\lesssim R_{\max}^2\bE_{\bD,\bT,\bZ}\bE_{(S,A)}\left[\TV^2\big(\cR(\cdot|  S,A),\widehat\cR(\cdot|  S,A)\big)\right]\\
&~~+Q_{\max}^2\bE_{\bD,\bT,\bZ}\bE_{(S,A)}\left[\TV^2\big(\cP(\cdot|  S,A),\widehat\cP(\cdot|  S,A)\big)\right]\\
&~~+\frac{\exp\left(\frac{Q_{\max}}{\alpha}\right)}{M}+\frac{Q_{\max}^2}{m}.
\end{align*}
Lemma \ref{lem:TV-flow-x} and Corollary
\ref{cor:TV-flow-r} complete the proof.
\end{proof}

\section{Proof of Theorem~\ref{thm:rl-error}}
\label{sec:value learning}

With the operator learning terms controlled, it remains to bound the value learning term in Lemma~\ref{lem:error-dec}. We first derive the H\"older regularity of $Q^*$ from the conditional density assumptions. We then show that the empirical soft optimal Bellman operator $\widehat{\mathcal{T}}_\alpha$ is a $\gamma$-contraction and control the statistical and ReLU approximation errors to prove Theorem~\ref{thm:rl-error}.

\begin{proof}
We write $\beta=k+\theta$, where $k\in\mathbb N_0$, and $\theta\in(0,1]$.For the optimal Gibbs policy, the soft continuation value is
\begin{align*}
V^*(x)
&=\mathbb E_{A'\sim\pi^*(\cdot|  x)}
\big[Q^*(x,A')-\alpha\log\pi^*(A'|  x)\big]\\
&=\alpha\log\int_{\cA}
\exp \left(\frac{Q^*(x,a')}{\alpha}\right) da'.
\end{align*}
Since $\|Q^*\|_\infty\le Q_{\max}$ and $|\cA|=1$,
\[
e^{-Q_{\max}/\alpha}
\le\int_{\cA}e^{Q^*(x,a')/\alpha} da'
\le e^{Q_{\max}/\alpha},
\]
Taking logarithms and multiplying by $\alpha$ gives
$-Q_{\max}\le V^*(x)\le Q_{\max}$ for every $x\in\cS$, and hence
$\|V^*\|_\infty\le Q_{\max}$. The soft optimal Bellman equation becomes
\begin{equation}\label{eq:soft-q-density-integral}
Q^*(w)=\int_0^{R_{\max}}r p_R(r|  w) dr
+\gamma\int_{\cS}V^*(x) p_1(x|  w) dx.
\end{equation}
Here $r$ and $x$ are integration variables. Differentiation in $w=(s,a)$ acts on the densities; $V^*(x)$ is a fixed bounded weight, and no differentiability of $V^*$ is assumed.

Assumption~\ref{ass:terminal-holder} gives, for every $|\nu|\le k$,
\[
|D_w^\nu p_R(r|  w)|\le L_R,
~~|D_w^\nu p_1(x|  w)|\le L_P,
\]
uniformly over $r\in[0,R_{\max}]$, $x\in\cS$, and $w\in\cS\times\cA$. For $|\nu|=k$ and $w,\widetilde w\in\cS\times\cA$, it also gives
\begin{align*}
|D_w^\nu p_R(r|  w)-D_w^\nu p_R(r| \widetilde w)|
&\le L_R\|w-\widetilde w\|^\theta,\\
|D_w^\nu p_1(x|  w)-D_w^\nu p_1(x| \widetilde w)|
&\le L_P\|w-\widetilde w\|^\theta.
\end{align*}
For $|\nu|<k$, let $e_\ell$ be the $\ell$th coordinate vector. At interior points $w,w+he_\ell$, the fundamental theorem of calculus gives
\[
\frac{D_w^\nu p_R(r|  w+he_\ell)-D_w^\nu p_R(r|  w)}{h}=\int_0^1D_w^{\nu+e_\ell}p_R(r|  w+uhe_\ell) du.
\]
This difference quotient is bounded in absolute value by $L_R$. The same identity for $p_1(x|  w)$ gives the bound $L_P$. Thus the weighted difference quotients in \eqref{eq:soft-q-density-integral} and its successive derivatives are dominated by $rL_R$ and $\gamma Q_{\max}L_P$, respectively. These bounds are independent of $h$ and integrable, since
\[
\int_0^{R_{\max}}rL_R dr=\frac{R_{\max}^2L_R}{2}<\infty,
~~
\int_{\cS}\gamma Q_{\max}L_P dx=\gamma Q_{\max}L_P<\infty.
\]
Repeated application of dominated convergence to the difference quotients yields, for every $|\nu|\le k$,
\begin{equation}\label{eq:soft-q-density-derivatives}
D_w^\nu Q^*(w)=\int_0^{R_{\max}}r D_w^\nu p_R(r|  w) dr
+\gamma\int_{\cS}V^*(x) D_w^\nu p_1(x|  w) dx.
\end{equation}
When $k=0$, this is simply the Bellman equation and no differentiation is needed.

Because $\cS$ has Lebesgue measure one, the derivative formula gives
\[
|D_w^\nu Q^*(w)|\le\frac{R_{\max}^2L_R}{2}
+\gamma L_P\|V^*\|_\infty
\le\frac{R_{\max}^2L_R}{2}+\gamma Q_{\max}L_P=:C_Q.
\]
For $|\nu|=k$, subtracting \eqref{eq:soft-q-density-derivatives} at two interior points gives
\begin{align*}
|D_w^\nu Q^*(w)-D_w^\nu Q^*(\widetilde w)|
&\le\int_0^{R_{\max}}r |D_w^\nu p_R(r|  w)-D_w^\nu p_R(r| \widetilde w)| dr\\
&~~+\gamma\int_{\cS}|V^*(x)| |D_w^\nu p_1(x|  w)-D_w^\nu p_1(x| \widetilde w)| dx\\
&\le C_Q\|w-\widetilde w\|^\theta.
\end{align*}
This also covers $k=0$. The density derivatives are continuous up to the boundary, so dominated convergence in \eqref{eq:soft-q-density-derivatives} extends the derivatives continuously to the closed state-action cube. The derivative and H\"older bounds hold there as well.

Finally, take H\"older extensions $\widetilde p_R,\widetilde p_1$ of the conditional densities with bounds $L_R,L_P$, and define, for $w\in\mathbb R^d$,
\[
\widetilde Q(w):=\int_0^{R_{\max}}r \widetilde p_R(r,w) dr
+\gamma\int_{\cS}V^*(x) \widetilde p_1(x,w) dx.
\]
The same estimates on $\mathbb R^d$ give $\widetilde Q\in\cH^\beta(\mathbb R^d)$. Since $\widetilde Q|_{\cS\times\cA}=Q^*$, we conclude that $Q^*\in\cH^\beta(\cS\times\cA)$.
\end{proof}

We next establish the contraction property used in the statistical error analysis.

\begin{lemma}\label{lem:T-widehat-gamma}
The operator $\widehat{\cT}_\alpha$ is a $\gamma$-contraction with respect to the sup-norm, that is, for any $Q_1,Q_2\in\mathcal Q$,
\[
\|\widehat{\cT}_\alpha Q_1-\widehat{\cT}_\alpha Q_2\|_\infty\le \gamma\|Q_1-Q_2\|_\infty.
\]
\end{lemma}

\begin{proof}
For any $Q_1,Q_2\in\mathcal Q$, define, for $k=1,2$,
\[
\widehat{V}_k(s):=\alpha\log\left(\frac1M\sum_{j=1}^M
\frac{\exp\left(\frac{1}{\alpha}Q_k(s,\tilde A_j)\right)}{f(\tilde A_j)}\right).
\]
Since
\[
Q_1(s,\tilde A_j)-Q_2(s,\tilde A_j)\le \|Q_1-Q_2\|_\infty,~~ f(\tilde A_j)>0,
\]
we have
\[
\begin{aligned}
\widehat{V}_1(s)-\widehat{V}_2(s)
&=\alpha\log\left(\frac{\frac1M\sum_{j=1}^M\frac{\exp\left(\frac{1}{\alpha}Q_1(s,\tilde A_j)\right)}{f(\tilde A_j)}}{\frac1M\sum_{j=1}^M\frac{\exp\left(\frac{1}{\alpha}Q_2(s,\tilde A_j)\right)}{f(\tilde A_j)}}\right)\\
&=\alpha\log\left(\frac{\frac1M\sum_{j=1}^M\frac{\exp\left(\frac{1}{\alpha}Q_2(s,\tilde A_j)\right)\exp\left(\frac{1}{\alpha}(Q_1(s,\tilde A_j)-Q_2(s,\tilde A_j))\right)}{f(\tilde A_j)}}{\frac1M\sum_{j=1}^M\frac{\exp\left(\frac{1}{\alpha}Q_2(s,\tilde A_j)\right)}{f(\tilde A_j)}}\right)\\
&\le \alpha\log\left(\frac{\exp\left(\frac{1}{\alpha}\|Q_1-Q_2\|_\infty\right)\frac1M\sum_{j=1}^M\frac{\exp\left(\frac{1}{\alpha}Q_2(s,\tilde A_j)\right)}{f(\tilde A_j)}}{\frac1M\sum_{j=1}^M\frac{\exp\left(\frac{1}{\alpha}Q_2(s,\tilde A_j)\right)}{f(\tilde A_j)}}\right)\\
&=\|Q_1-Q_2\|_\infty.
\end{aligned}
\]
Exchanging $Q_1$ and $Q_2$ gives
\[
\widehat{V}_2(s)-\widehat{V}_1(s)\le \|Q_1-Q_2\|_\infty.
\]
Thus,
\[
|\widehat{V}_1(s)-\widehat{V}_2(s)|\le \|Q_1-Q_2\|_\infty.
\]
Therefore, for any $(s,a)$,
\[
\begin{aligned}
|\widehat{\cT}_\alpha Q_1(s,a)-\widehat{\cT}_\alpha Q_2(s,a)|
&=\left|\frac{\gamma}{m}\sum_{i=1}^m\left(\widehat{V}_1(s_i')-\widehat{V}_2(s_i')\right)\right|\\
&\le \frac{\gamma}{m}\sum_{i=1}^m|\widehat{V}_1(s_i')-\widehat{V}_2(s_i')|\\
&\le \gamma\|Q_1-Q_2\|_\infty.
\end{aligned}
\]
Taking the supremum over $(s,a)$ gives
\[
\|\widehat{\cT}_\alpha Q_1-\widehat{\cT}_\alpha Q_2\|_\infty\le \gamma\|Q_1-Q_2\|_\infty.
\]
This completes the proof.
\end{proof}

The contraction property provides the stability needed to compare Bellman residual losses uniformly over $\mathcal Q$. We now use it together with a covering number argument to control the statistical error of empirical residual minimization.

\subsection{Proof of Lemma \ref{lem:term-2-statistical}} \label{sec:apd-term-2-statistical}

\begin{proof}[Proof of Lemma \ref{lem:term-2-statistical}]
All probabilities and expectations below are conditional on the fixed learned operator, which is independent of $\mathbb S$. For each $Q\in\mathcal Q$, define
\[
D_Q:=\widehat{\mathcal T}_\alpha Q-Q,~~\ell_Q(s,a):=|D_Q(s,a)|^2.
\]
Assume that every generated reward lies in $[0,R_{\max}]$ and every generated next state lies in $\mathcal S$. For this proof, set
\[
B:=2Q_{\max}+\gamma\alpha\max\{|\log c_1|,|\log c_2|\}.
\]
The parameters $Q_{\max}>0$, $\alpha$, $\gamma$, $c_1$, and $c_2$ are fixed with respect to $n$; the implicit constants may depend on these parameters, so $B\lesssim Q_{\max}$. We first verify that
\[
0\le\ell_Q(s,a)\le\ell_{\max}:=B^2,~~Q\in\mathcal Q,~~(s,a)\in\mathcal S\times\mathcal A.
\]
By monotonicity of the logarithm,
\[
\log c_1\le\log f(a)\le\log c_2,
~~|\log f(a)|\le \max\{|\log c_1|,|\log c_2|\},~~ a\in\mathcal A.
\]
The bounds on $Q$ and $f$, followed by averaging the positive importance weights and applying $\alpha\log$, give
\[
-Q_{\max}-\alpha\log c_2\le\widehat V_Q(s')\le Q_{\max}-\alpha\log c_1,~~s'\in\mathcal S.
\]
Hence $\|\widehat V_Q\|_\infty\le Q_{\max}+\alpha \max\{|\log c_1|,|\log c_2|\}$. Consequently,
\[
\begin{aligned}
\|D_Q\|_\infty
&\le R_{\max}+(1+\gamma)Q_{\max}+\gamma\alpha \max\{|\log c_1|,|\log c_2|\}\\
&\le 2Q_{\max}+\gamma\alpha \max\{|\log c_1|,|\log c_2|\}=B,
\end{aligned}
\]
where the last inequality uses $R_{\max}\le(1-\gamma)Q_{\max}$.
For $Q_1,Q_2\in\mathcal Q$, the difference-of-squares identity and the contraction property give
\[
\begin{aligned}
\|\ell_{Q_1}-\ell_{Q_2}\|_\infty
&\le\|D_{Q_1}+D_{Q_2}\|_\infty
\|D_{Q_1}-D_{Q_2}\|_\infty\\
&\le2B(1+\gamma)\|Q_1-Q_2\|_\infty.
\end{aligned}
\]
Set
\[
N:=\mathcal N\left(\mathcal Q,\frac{B}{2(1+\gamma)n},\|\cdot\|_\infty\right),
\]
and let $\{Q_1,\ldots,Q_N\}$ be a corresponding cover of $\mathcal Q$. For every $Q\in\mathcal Q$, there exists $Q_{k^*}$ such that
\[
\|Q-Q_{k^*}\|_\infty\le\frac{B}{2(1+\gamma)n},
~
\|\ell_Q-\ell_{Q_{k^*}}\|_\infty\le\frac{\ell_{\max}}{n}.
\]
Consequently,
\[
\mathcal L(\widehat{\cT}_\alpha;Q)-2\widehat{\mathcal L}(\widehat{\cT}_\alpha;Q)\le \mathcal L(\widehat{\cT}_\alpha;Q_{k^*})-2\widehat{\mathcal L}(\widehat{\cT}_\alpha;Q_{k^*})+\frac{3\ell_{\max}}{n}.
\]
Therefore,
\[
\sup_{Q\in\mathcal Q}\Big(\mathcal L(\widehat{\cT}_\alpha;Q)-2\widehat{\mathcal L}(\widehat{\cT}_\alpha;Q)\Big)
\le\max_{1\le k\le N}\Big(\mathcal L(\widehat{\cT}_\alpha;Q_k)-2\widehat{\mathcal L}(\widehat{\cT}_\alpha;Q_k)\Big)+\frac{3\ell_{\max}}{n}.
\]
Fix $Q\in\mathcal Q$. For every $\delta\in(0,1)$, Chernoff's inequality with parameter $\frac{3n}{7\ell_{\max}}$ gives
\begin{align*}
  &\mathbb{P}\left(\mathcal{L}({\widehat{\mathcal{T}}}_{\alpha};Q)-2\widehat{\mathcal{L}}({\widehat{\mathcal{T}}}_{\alpha};Q)>\frac{7\ell_{\max}}{3n}\log\frac{N}{\delta}\right)\\
  &\le
  \exp\Big(-\log\frac{N}{\delta}\Big)\mathbb{E}\exp\Big(\frac{3n}{7\ell_{\max}}\big(\mathcal{L}({\widehat{\mathcal{T}}}_{\alpha};Q)-2\widehat{\mathcal{L}}({\widehat{\mathcal{T}}}_{\alpha};Q)\big)\Big).
\end{align*}
It remains to show that the exponential moment is bounded by one.  Since
the squared Bellman residual is bounded between \(0\) and \(\ell_{\max}\), the convexity of \(x\mapsto e^{-\frac{6x}{7\ell_{\max}}}\) on
\([0,\ell_{\max}]\) implies
\[
e^{-\frac{6x}{7\ell_{\max}}}\le 1-\frac{1-e^{-\frac{6}{7}}}{\ell_{\max}}x,~~ 0\le x\le \ell_{\max} .
\]
It follows that
\[
\mathbb E\exp\left(-\frac{6\ell_Q(S,A)}{7\ell_{\max}}\right)
\le\exp\left(-\frac{1-\exp(-6/7)}{\ell_{\max}}\mathcal L(\widehat{\cT}_\alpha;Q)\right).
\]
Using independence,
\begin{align*}
  \mathbb{E}\exp\Big(\frac{3n}{7\ell_{\max}}\left(\mathcal{L}({\widehat{\mathcal{T}}}_{\alpha};Q)-2\widehat{\mathcal{L}}({\widehat{\mathcal{T}}}_{\alpha};Q)\right)\Big) 
  & =
  \exp\Big({\frac{3n}{7\ell_{\max}}\mathcal{L}({\widehat{\mathcal{T}}}_{\alpha};Q)}\Big)\\
  & ~~ \times
  \left(\mathbb{E} \exp\Big({-6\left|{\widehat{\mathcal{T}}}_{\alpha}Q-Q\right|^2/(7\ell_{\max})\Big)}\right)^n \\
  &\le
  \exp\Big(\left[\frac{3n}{7\ell_{\max}}-\frac{n}{\ell_{\max}}\left(1-e^{-\frac{6}{7}}\right)\right]\mathcal{L}({\widehat{\mathcal{T}}}_{\alpha};Q)\Big).
\end{align*}
Moreover,
$  1-e^{-\frac{6}{7}}\ge \frac37 ,$
and consequently
$  \frac{3n}{7\ell_{\max}}-\frac{n}{\ell_{\max}}\left(1-e^{-\frac{6}{7}}\right)\le0 .$
Hence the exponential moment is at most one, and so
\[
\mathbb{P}\left(\mathcal{L}({\widehat{\mathcal{T}}}_{\alpha};Q)-2\widehat{\mathcal{L}}({\widehat{\mathcal{T}}}_{\alpha};Q)>\frac{7\ell_{\max}}{3n}\log\frac{N}{\delta}\right)\le \frac{\delta}{N}.
\]
Taking a union bound over the elements of the net yields
\[
  \mathbb{P}\left(\sup_{Q\in\mathcal{Q}}\Big(\mathcal{L}({\widehat{\mathcal{T}}}_{\alpha};Q)-2\widehat{\mathcal{L}}({\widehat{\mathcal{T}}}_{\alpha};Q)\Big)>\frac{3\ell_{\max}}{n}+\frac{7\ell_{\max}}{3n}\log\frac{N}{\delta}\right)
  \le \delta .
\]
Equivalently, for all
$  t\ge\frac{3\ell_{\max}}{n}+\frac{7\ell_{\max}}{3n}\log N,$
we have
\[
  \mathbb{P}\left(\sup_{Q\in\mathcal{Q}}\Big(\mathcal{L}({\widehat{\mathcal{T}}}_{\alpha};Q)-2\widehat{\mathcal{L}}({\widehat{\mathcal{T}}}_{\alpha};Q)\Big)>t\right)
  \le
  N\exp\Big(-\frac{3n(t-\frac{3\ell_{\max}}{n})}{7\ell_{\max}}\Big).
\]
Set $a_n:=\frac{3\ell_{\max}}{n}+\frac{7\ell_{\max}}{3n}\log N$. Using the positive part tail integral identity, we obtain
\begin{align*}
  \mathbb{E}_\mathbb{S}\sup_{Q\in\mathcal{Q}}\Big(\mathcal{L}({\widehat{\mathcal{T}}}_{\alpha};Q)-2\widehat{\mathcal{L}}({\widehat{\mathcal{T}}}_{\alpha};Q)\Big)
  &\le\int_0^\infty\mathbb{P}\left(\sup_{Q\in\mathcal{Q}}\Big(\mathcal{L}({\widehat{\mathcal{T}}}_{\alpha};Q)-2\widehat{\mathcal{L}}({\widehat{\mathcal{T}}}_{\alpha};Q)\Big)>t\right) dt \\
  &=\int_0^{a_n}\mathbb{P}\left(\sup_{Q\in\mathcal{Q}}\Big(\mathcal{L}({\widehat{\mathcal{T}}}_{\alpha};Q)-2\widehat{\mathcal{L}}({\widehat{\mathcal{T}}}_{\alpha};Q)\Big)>t\right) dt \\
  &+\int_{a_n}^{\infty}
  \mathbb{P}\left(\sup_{Q\in\mathcal{Q}}\Big(\mathcal{L}({\widehat{\mathcal{T}}}_{\alpha};Q)-2\widehat{\mathcal{L}}({\widehat{\mathcal{T}}}_{\alpha};Q)\Big)>t\right) dt \\
  &\le a_n +\frac{7\ell_{\max}}{3n}.
\end{align*}
The tail integral used above is
\[
\int_{a_n}^{\infty}N\exp\Big(-\frac{3n(t-\frac{3\ell_{\max}}{n})}{7\ell_{\max}}\Big) dt=\frac{7\ell_{\max}}{3n}.
\]
Hence
\[
  \mathbb{E}_{\bS}\sup_{Q\in\mathcal{Q}}\Big(\mathcal{L}({\widehat{\mathcal{T}}}_{\alpha};Q)-2\widehat{\mathcal{L}}({\widehat{\mathcal{T}}}_{\alpha};Q)\Big)
  \le
  \frac{3\ell_{\max}}{n}+\frac{7\ell_{\max}}{3n}\Big(\log N+1\Big).
\]
Since $B$ and $\gamma$ are fixed, Lemma~\ref{lem:cover-nn} gives
\[
\begin{aligned}
\log N
&\lesssim\mathcal C\mathcal D
\log\left(1+\frac{2(1+\gamma)n\mathcal B\mathcal W\mathcal D}{B}\right)\\
&\lesssim n^{\frac{d}{2\beta+d}}\log^3n.
\end{aligned}
\]
Substituting this covering bound and $\ell_{\max}=B^2\lesssim Q_{\max}^2$ into the preceding expectation inequality yields
\[
\begin{aligned}
&\mathbb E_{\mathbb S}\sup_{Q\in\mathcal Q}
\bigl(\mathcal L(\widehat{\mathcal T}_\alpha;Q)
-2\widehat{\mathcal L}(\widehat{\mathcal T}_\alpha;Q)\bigr)\\
&~~\lesssim Q_{\max}^2n^{-\frac{2\beta}{2\beta+d}}\log^3n.
\end{aligned}
\]
This completes the proof.
\end{proof}

The preceding lemma controls the statistical component of value learning. Combining it with Lemma~\ref{lem:soft-q-holder} and the ReLU approximation bound for $Q^*$ yields the complete value learning guarantee.

\subsection{Proof of Theorem \ref{thm:rl-error}}\label{sec:apd-term-2}
\begin{proof}[Proof of Theorem \ref{thm:rl-error}]
By Lemma~\ref{lem:soft-q-holder}, Assumption~\ref{ass:terminal-holder} implies that $Q^*/Q_{\max}\in\cH^\beta$ with a fixed H\"older radius bounded by $2C_Q/Q_{\max}$. Choose the ReLU network class $\cQ$ with network size $\mathcal C=\cO\bigl(n^{\frac{d}{d+2\beta}}\log n\bigr)$, depth $\mathcal D=\cO(\log n)$, width $\mathcal W=\cO\bigl(n^{\frac{d}{d+2\beta}}\bigr)$, and parameter bound $\mathcal B=\cO\bigl(n^{\frac{d}{d+2\beta}}\bigr)$. Apply Lemma~\ref{lem:ApproximationError} to $Q^*/Q_{\max}$ with approximation tolerance $\varepsilon=n^{-\frac{\beta}{d+2\beta}}$ and rescale the network output by $Q_{\max}$. Since $Q_{\max}$ is fixed, the rescaled network has the same complexity orders, and
\begin{equation}\label{eq:RL-approximation-error}
\inf_{Q\in\cQ}\|Q-Q^*\|_\infty^2\lesssim Q_{\max}^2 n^{-\frac{2\beta}{d+2\beta}}.
\end{equation}
Combining \eqref{eq:RL-approximation-error} with the statistical error bound in Lemma~\ref{lem:term-2-statistical} establishes the claimed value learning error bound and completes the proof.
\end{proof}

\section{Proof of Theorem~\ref{thm:main-result}}
\label{sec:main-results}

We first establish the loss comparison and uniqueness result used in Section~\ref{method} to characterize the population objective. We then derive the error decomposition that connects this objective to the finite-sample estimator $\widehat Q$.

\begin{lemma}\label{lem:soft-residual-transfer}
Let $\mu=\mu_{\pi_b}$ and $\mu_{\pi^*}$ be the state action distributions introduced in Section~\ref{sec:entropy-mdp}. If $\mu_{\pi^*}\ll\mu$ and $C_\mu:=\left\|\frac{d\mu_{\pi^*}}{d\mu}\right\|_{L^2(\mu)}<\infty$, then, for every continuous function $Q$ satisfying $\|Q\|_\infty\le Q_{\max}$,
\[
\bigl[\mathcal L_{\pi^*}(\mathcal T_\alpha^*;Q)\bigr]^2\le4C_\mu^2Q_{\max}^2 \mathcal L(\mathcal T_\alpha^*;Q).
\]
Furthermore, under Assumption~\ref{ass:terminal-holder} and the common full-support condition in Section~\ref{sec:entropy-mdp}, both losses have the same unique minimizer among continuous functions $Q$ satisfying $\|Q\|_\infty\le Q_{\max}$:
\[
\arg\min_Q\mathcal L(\mathcal T_\alpha^*;Q)=\arg\min_Q\mathcal L_{\pi^*}(\mathcal T_\alpha^*;Q)=\{Q^*\}.
\]
\end{lemma}

\begin{proof}
For $\|Q\|_\infty\le Q_{\max}$, we recall that
\[
V_Q(s)=\alpha\log\int_{\mathcal A}\exp\left(\frac{Q(s,a)}{\alpha}\right) da.
\]
Since $|\mathcal A|=1$ and $\|Q\|_\infty\le Q_{\max}$, we have $\|V_Q\|_\infty\le Q_{\max}$.
Writing $\Delta_Q:=\mathcal T_\alpha^*Q-Q$ and using
$R_{\max}\le(1-\gamma)Q_{\max}$, we obtain
\[
\|\Delta_Q\|_\infty\le R_{\max}+(1+\gamma)Q_{\max}\le2Q_{\max}.
\]
A change of measure followed by the Cauchy--Schwarz inequality therefore gives
\begin{align*}
\bigl[\mathcal L_{\pi^*}(\mathcal T_\alpha^*;Q)\bigr]^2
&=\left(\int_{\mathcal S\times\mathcal A}\frac{d\mu_{\pi^*}}{d\mu} |\Delta_Q|^2 d\mu\right)^2 \le C_\mu^2\int_{\mathcal S\times\mathcal A}|\Delta_Q|^4 d\mu\\
&\le C_\mu^2\|\Delta_Q\|_\infty^2\mathcal L(\mathcal T_\alpha^*;Q)\le4C_\mu^2Q_{\max}^2 \mathcal L(\mathcal T_\alpha^*;Q).
\end{align*}
Next, we prove uniqueness. By Lemma~\ref{lem:soft-q-holder}, Assumption~\ref{ass:terminal-holder} guarantees that $Q^*$ is continuous. Together with $\|Q^*\|_\infty\le Q_{\max}$ and $\mathcal T_\alpha^*Q^*=Q^*$, this shows that both losses attain their global minimum value zero at $Q^*$. For any continuous $Q$ on $\mathcal S\times\mathcal A$ with $\|Q\|_\infty\le Q_{\max}$, the soft Bellman integral representation is
\[
\mathcal T_\alpha^*Q(s,a)=\int_0^{R_{\max}}r p_R(r|  s,a) dr+\gamma\int_{\mathcal S}V_Q(x)p_1(x|  s,a) dx.
\]
Assumption~\ref{ass:terminal-holder} ensures that the conditional densities are continuous in $(s,a)$ and uniformly bounded on their compact domains. Together with the boundedness of $V_Q$, these properties allow us to apply dominated convergence to both integrals. Thus $\mathcal T_\alpha^*Q$ is continuous, and so is $\Delta_Q$.

Let $Q$ be any global minimizer of
$\mathcal L(\mathcal T_\alpha^*;\cdot)$ among continuous functions with $\|Q\|_\infty\le Q_{\max}$. Then
$\mathcal L(\mathcal T_\alpha^*;Q)=0$.
If $\Delta_Q$ were nonzero at some point, its continuity would yield a nonempty relatively open neighborhood $U\subseteq\mathcal S\times\mathcal A$ and $\varepsilon>0$ such that $|\Delta_Q|\ge\varepsilon$ on $U$. The full-support condition gives $\mu(U)>0$, and hence
\[
\mathcal L(\mathcal T_\alpha^*;Q)\ge\varepsilon^2\mu(U)>0,
\]
a contradiction. Consequently, $Q=\mathcal T_\alpha^*Q$ throughout $\mathcal S\times\mathcal A$. By the $\gamma$-contraction property of $\mathcal T_\alpha^*$,
\[
\|Q-Q^*\|_\infty=\|\mathcal T_\alpha^*Q-\mathcal T_\alpha^*Q^*\|_\infty\le\gamma\|Q-Q^*\|_\infty.
\]
Since $\gamma<1$, this implies $Q=Q^*$. Repeating the same argument with $\mu_{\pi^*}$ in place of $\mu$ establishes uniqueness for
$\mathcal L_{\pi^*}(\mathcal T_\alpha^*;\cdot)$.
\end{proof}

\subsection{Proof of Lemma~\ref{lem:error-dec}}
\begin{proof}
Condition on $\bD,\bT,\bZ$. Since
$\mathcal L(\mathcal T_\alpha^*;Q^*)=\|\cT^*_\alpha Q^*-Q^*\|_{L^2(\mu)}^2=0$,
we have
\[
\begin{aligned}
\mathcal L(\mathcal T_\alpha^*;\widehat Q)-\mathcal L(\mathcal T_\alpha^*;Q^*)
&=\|\cT^*_\alpha\widehat Q-\widehat{\cT}_\alpha\widehat Q+\widehat{\cT}_\alpha\widehat Q-\widehat Q\|_{L^2(\mu)}^2\\
&\le 2\|\cT^*_\alpha\widehat Q-\widehat{\cT}_\alpha\widehat Q\|_{L^2(\mu)}^2+2\cL(\widehat{\cT}_\alpha;\widehat Q).
\end{aligned}
\]
Since
\[
\begin{aligned}
2\cL(\widehat{\cT}_\alpha;\widehat Q)
&=2\Big(\cL(\widehat{\cT}_\alpha;\widehat Q)-2\widehat{\cL}(\widehat{\cT}_\alpha;\widehat Q)\Big)+4\widehat{\cL}(\widehat{\cT}_\alpha;\widehat Q)\\
&\le 2\sup_{Q\in\mathcal Q}\Big(\cL(\widehat{\cT}_\alpha;Q)-2\widehat{\cL}(\widehat{\cT}_\alpha;Q)\Big)+4\inf_{Q\in\mathcal Q}\widehat{\cL}(\widehat{\cT}_\alpha;Q),
\end{aligned}
\]
then
\[
\begin{aligned}
\mathcal L(\mathcal T_\alpha^*;\widehat Q)-\mathcal L(\mathcal T_\alpha^*;Q^*)
&\le 2\|\cT^*_\alpha\widehat Q-\widehat{\cT}_\alpha\widehat Q\|_{L^2(\mu)}^2+2\sup_{Q\in\mathcal Q}\Big(\cL(\widehat{\cT}_\alpha;Q)-2\widehat{\cL}(\widehat{\cT}_\alpha;Q)\Big)\\
&~~ +4\inf_{Q\in\mathcal Q}\widehat{\cL}(\widehat{\cT}_\alpha;Q).
\end{aligned}
\]
Taking expectation over $\bS$ yields
\[
\begin{aligned}
\bE_{\bS}\left[\mathcal L(\mathcal T_\alpha^*;\widehat Q)-\mathcal L(\mathcal T_\alpha^*;Q^*)\right]
&\le 2\bE_{\bS}\left[\|\cT^*_\alpha\widehat Q-\widehat{\cT}_\alpha\widehat Q\|_{L^2(\mu)}^2\right] +2\bE_{\bS}\sup_{Q\in\mathcal Q}\Big(\cL(\widehat{\cT}_\alpha;Q)-2\widehat{\cL}(\widehat{\cT}_\alpha;Q)\Big)\\
&~~ +4\bE_{\bS}\inf_{Q\in\mathcal Q}\widehat{\cL}(\widehat{\cT}_\alpha;Q).
\end{aligned}
\]
Since $\bS$ is independent of $\widehat{\cT}_\alpha$ conditional on $\bD,\bT,\bZ$,
\[
\bE_{\bS}\inf_{Q\in\mathcal Q}\widehat{\cL}(\widehat{\cT}_\alpha;Q)\le \inf_{Q\in\mathcal Q}\bE_{\bS}\widehat{\cL}(\widehat{\cT}_\alpha;Q)=\inf_{Q\in\mathcal Q}\cL(\widehat{\cT}_\alpha;Q).
\]
For any $Q\in\mathcal Q$, using $\cT^*_\alpha Q^*=Q^*$,
\[
\widehat{\cT}_\alpha Q-Q=\widehat{\cT}_\alpha Q-\widehat{\cT}_\alpha Q^*+\widehat{\cT}_\alpha Q^*-\cT^*_\alpha Q^*+Q^*-Q.
\]
Thus,
\[
\begin{aligned}
\cL(\widehat{\cT}_\alpha;Q)
&=\left\|\widehat{\cT}_\alpha Q-\widehat{\cT}_\alpha Q^*+\widehat{\cT}_\alpha Q^*-\cT^*_\alpha Q^*+Q^*-Q\right\|_{L^2(\mu)}^2\\
&\le 2\|\widehat{\cT}_\alpha Q^*-\cT^*_\alpha Q^*\|_{L^2(\mu)}^2+2\left(\|\widehat{\cT}_\alpha Q-\widehat{\cT}_\alpha Q^*\|_{L^2(\mu)}+\|Q-Q^*\|_{L^2(\mu)}\right)^2\\
&\le 2\|\widehat{\cT}_\alpha Q^*-\cT^*_\alpha Q^*\|_{L^2(\mu)}^2+2(1+\gamma)^2\|Q-Q^*\|_\infty^2.
\end{aligned}
\]
Here,
\[
\|\widehat{\cT}_\alpha Q-\widehat{\cT}_\alpha Q^*\|_{L^2(\mu)}\le \|\widehat{\cT}_\alpha Q-\widehat{\cT}_\alpha Q^*\|_\infty\le \gamma\|Q-Q^*\|_\infty,
\]
and
\[
\|Q-Q^*\|_{L^2(\mu)}\le \|Q-Q^*\|_\infty.
\]
Taking the infimum over $Q\in\mathcal Q$ gives
\[
4\inf_{Q\in\mathcal Q}\cL(\widehat{\cT}_\alpha;Q)\le 8\|\widehat{\cT}_\alpha Q^*-\cT^*_\alpha Q^*\|_{L^2(\mu)}^2+8(1+\gamma)^2\inf_{Q\in\mathcal Q}\|Q-Q^*\|_\infty^2.
\]
Substituting this bound back, we obtain
\[
\begin{aligned}
\bE_{\bS}\left[\mathcal L(\mathcal T_\alpha^*;\widehat Q)-\mathcal L(\mathcal T_\alpha^*;Q^*)\right]
&\le 2\bE_{\bS}\left[\|\cT^*_\alpha\widehat Q-\widehat{\cT}_\alpha\widehat Q\|_{L^2(\mu)}^2\right]+8\|\widehat{\cT}_\alpha Q^*-\cT^*_\alpha Q^*\|_{L^2(\mu)}^2\\
&~~ +2\bE_{\bS}\sup_{Q\in\mathcal Q}\Big(\cL(\widehat{\cT}_\alpha;Q)-2\widehat{\cL}(\widehat{\cT}_\alpha;Q)\Big)\\
&~~ +8(1+\gamma)^2\inf_{Q\in\mathcal Q}\|Q-Q^*\|_\infty^2.
\end{aligned}
\]
This completes the proof.
\end{proof}

The error decomposition established in Appendix~\ref{sec:main-results} separates the excess Bellman residual risk into operator learning and value learning contributions. Combining this decomposition with the bounds proved in Appendices~\ref{sec:soft-bellman} and~\ref{sec:value learning} yields Theorem~\ref{thm:main-result}.

\begin{proof}[Proof of Theorem \ref{thm:main-result}]
By the error decomposition of Lemma \ref{lem:error-dec}, together with Theorem \ref{thmm:rl-flow-bound}--\ref{thm:rl-error}, the desired theorem follows directly.
\end{proof}

\section{Auxiliary results}\label{sec:apd-auxiliary-results}
For completeness, we record the neural network approximation and covering results used in the value learning and operator learning arguments above.

\begin{lemma}[Lemma 6.1 of \citet{jiao2025deep}]
\label{lem:ApproximationError}
Assume that $f \in \mathcal{H}^{\varsigma}$ with $\varsigma = s + r$,
$s \in \mathbb{N}_0$ and $r \in (0,1]$. For any $\varepsilon \in (0,1)$,
there exists a ReLU DNN function $\psi$ with depth \(\cD \leq \cO\bigl(\log(1/\varepsilon)\bigr)\),
size \(\mathcal C \leq\cO\bigl(\varepsilon^{-d/\varsigma}\log(1/\varepsilon)\bigr),\)
and weight bound \(\cB \leq \cO\bigl(\varepsilon^{-d/\varsigma}\bigr)\)
such that
\[
\|f-\psi\|_{\infty} \leq \varepsilon.
\]
\end{lemma}

\begin{lemma}[Lemma 20 of \citet{feng2024deep}]
\label{lem:cover-nn}
Let $\mathcal{F}$ be the ReLU DNNs with width $\cW$, depth $\cD$, and size $\mathcal C$.
Assume that the parameters of $\mathcal{F}$ are bounded by a constant $\cB > 0$.
Then, for each $\delta > 0$,
\[
\log \mathcal{N}\bigl(\mathcal{F}, \delta, \|\cdot\|_{\infty}\bigr)\leq \cO(\mathcal C\cD)\log(\cB\cW\cD/\delta).
\]
\end{lemma}

\end{appendix}

%%%%%%%%%%%%%%%%%%%%%%%%%%%%%%%%%%%%%%%%%%%%%%%%
%%%%%%%%%%%%%%%%%%%%%%%%%%%%%%%%%%%%%%%%%%%%%%%%
\bibliographystyle{spbasic}   
\bibliography{ref}  

\end{document}